\documentclass[10pt,twocolumn,letterpaper]{article}

\usepackage{cvpr}              % To produce the CAMERA-READY version

\usepackage{gensymb}

\newcommand{\ourmodel}[1]{MAGICIAN\xspace}

\definecolor{cvprblue}{rgb}{0.21,0.49,0.74}
\PassOptionsToPackage{hyphens}{url}

\usepackage[pagebackref,breaklinks,colorlinks,allcolors=cvprblue]{hyperref}

\usepackage{dsfont}
\usepackage{etoolbox}
\usepackage{color}
\usepackage{tikz}
\usetikzlibrary{shadings}
\usepackage{multirow}
\usepackage{tikz}
\usepackage{tikz-3dplot}
\usetikzlibrary{arrows.meta, shapes, positioning, decorations.pathreplacing}
\tikzset{
  ball/.style={
    shading=radial,
    inner color=#1!100,
    outer color=#1!20
  }
}

\makeatletter
\DeclareRobustCommand*\cal{\@fontswitch\relax\mathcal}
\makeatother

\newif\ifshowedits

\newcommand{\addeditor}[3]{%
  \definecolor{#1color}{rgb}{#3}
  \expandafter\newcommand\csname #1\endcsname[1]{%
  \ifshowedits
    {\color{#1color} ##1}%
  \else
    {##1}%
  \fi
  }%
  \expandafter\newcommand\csname #1rmk\endcsname[1]{%
  \ifshowedits
    {\color{#1color} {\bf [#2: ##1]}}
  \fi
  }%
  \expandafter\newcommand\csname #1rpl\endcsname[2]{%
  \ifshowedits
    {\color{#1color} ##1 \sout{##2}}
  \else
    {##1}
  \fi
  }%
}

\newcommand{\createtextvar}[1]{
  \expandafter\newcommand\csname #1\endcsname{%
  {\text{#1}}
}%
}
\newcommand{\mycomment}[1]{}

\newcommand{\calE}{{\cal E}}

\newcommand{\bc}{{\bf c}}

\newcommand{\bx}{{\bf x}}

\newcommand{\bC}{{\bf C}}

\newcommand{\IR}{{\mathds{R}}}

\newcommand{\vcomment}[1]{}

\usepackage{colortbl}

\definecolor{yellow}{rgb}{1, 1, 0.7}
\definecolor{orange}{rgb}{1, 0.85, 0.7}
\definecolor{tablered}{rgb}{1, 0.7, 0.7}
\definecolor{red}{rgb}{1, 0, 0}

\definecolor{tablethree}{rgb}{0.7, 1, 1}
\definecolor{tabletwo}{rgb}{0.7, 0.85, 1}
\definecolor{tableone}{rgb}{0.7, 0.7, 1}

\newcommand{\best}{\cellcolor{tablered}}
\newcommand{\sbest}{\cellcolor{orange}}
\newcommand{\tbest}{\cellcolor{yellow}}

\newcommand{\VF}{{\text{VF}}}
\newcommand{\cam}{{\mathbf{c}}}
\newcommand{\vecx}{{\mathbf{x}}}

\newcommand{\occ}{{\hat{\sigma}}}

\newcommand{\nov}{{\hat{\gamma}}}
\newcommand{\pix}{\mathbf{p}}
\newcommand{\occl}{\hat{o}}
\newcommand{\vecd}{{\mathbf{d}}}
\newcommand{\veco}{{\mathbf{o}}}

\addeditor{shiyao}{SL}{0.8, 0.4, 0.0}
\addeditor{antoine}{AG}{0.0, 0.0, 0.75}
\addeditor{vincent}{VL}{0.0, 0.5, 0.0}
\addeditor{shizhe}{SC}{0.8, 0.0, 0.6}
\showeditstrue
\def\paperID{} % *** Enter the Paper ID here
\def\confName{CVPR}
\def\confYear{2026}

\title{
\ourmodel{}:
Efficient Long-Term Planning with \\Imagined Gaussians for Active Mapping 
\vspace{-0.3em}
}

\author{
Shiyao Li$^{1}$ \quad Antoine Guédon$^{1, 2}$ \quad Shizhe Chen$^{3}$ \quad Vincent Lepetit$^{1}$\\
$^{1}$LIGM, \'Ecole Nationale des Ponts et Chauss\'ees, IP Paris, Univ Gustave Eiffel, CNRS, France\\
$^{2}$\'Ecole Polytechnique, France\\
$^{3}$Inria, \'Ecole normale sup\'erieure, CNRS, PSL Research University, France\\
}
\begin{document}
\twocolumn[{
\renewcommand\twocolumn[1][]{#1}
\maketitle
\begin{center}
\vspace{-0.5em}
    \includegraphics[width=1.\textwidth]{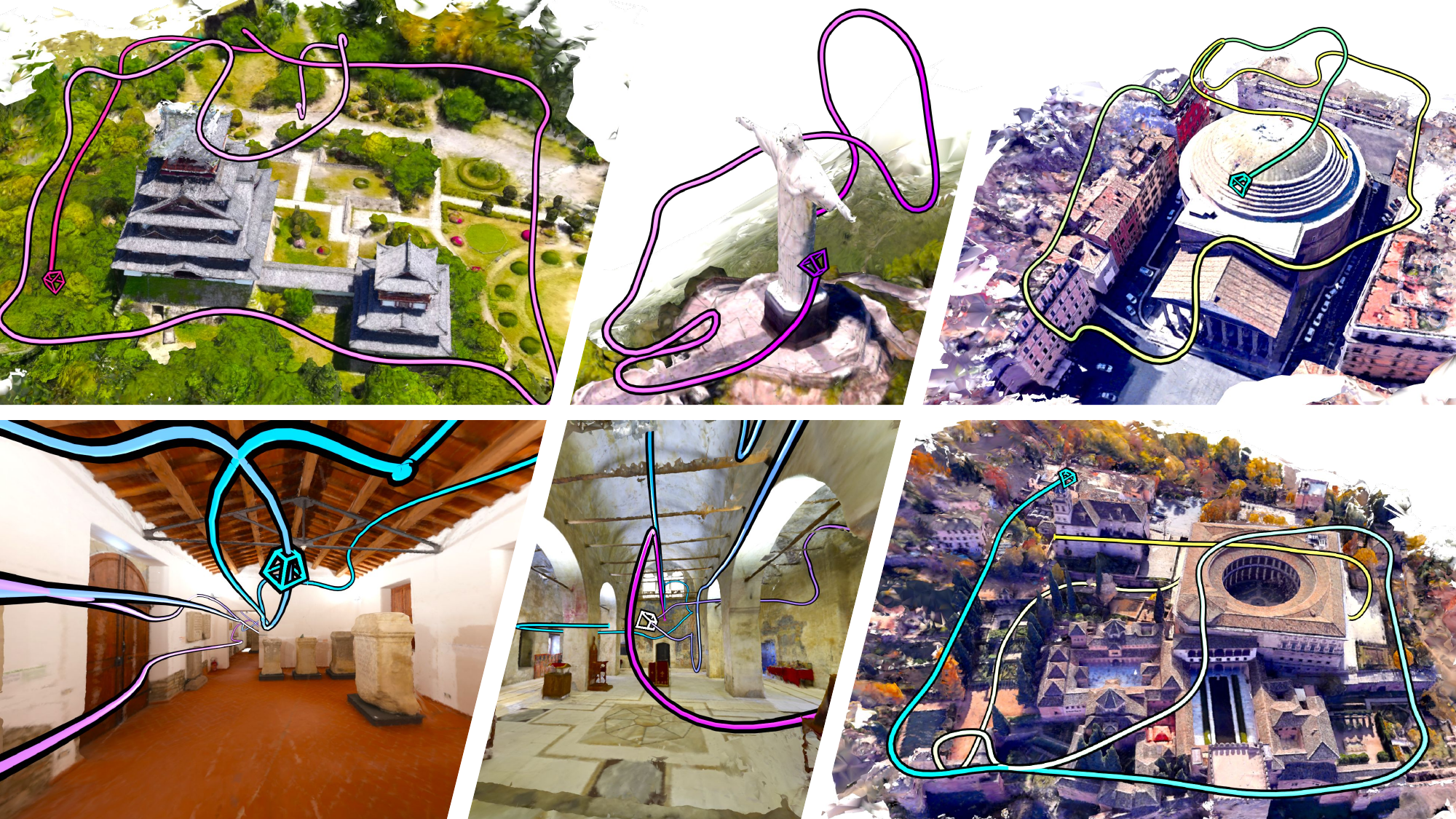}
    \captionof{figure}{
    \textbf{\ourmodel{} enables efficient, high-coverage exploration across diverse environments.}
    We visualize the exploration trajectories (light-to-dark gradients) generated by our method and the resulting 3D reconstructions (surface meshes and textures) for various outdoor and indoor scenes.
    \ourmodel{} is powered by what we call ``Imagined Gaussians'', predicted by our occupancy model to model scene uncertainty, making efficient long-term planning possible. 
    %The resulting trajectories     efficiently cover the scene and enable accurate geometry reconstruction and novel view synthesis.
    }
    \label{fig:teaser}
\end{center}
}]

\begin{abstract}
Active mapping aims to determine how an agent should move to efficiently reconstruct unknown environments. 
Most existing approaches rely on greedy next-best-view prediction, resulting in inefficient exploration and incomplete reconstruction.
To address this, we introduce \ourmodel{}, a novel long-term planning framework that maximizes accumulated surface coverage gain through \textit{Imagined Gaussians}, a scene representation based on 3D Gaussian Splatting, derived from a pre-trained occupancy network with strong structural priors. 
% This representation enables efficient computation of coverage gain for any novel viewpoint via fast volumetric rendering.
% The resulting speedup allows the integration of the gain metric into a tree-search algorithm for planning long-horizon paths.
This representation enables efficient coverage gain computation for any novel viewpoint via fast volumetric rendering, allowing its integration into a tree-search algorithm for long-horizon planning.
We update Imagined Gaussians and refine the trajectory in a closed loop.
Our method achieves state-of-the-art performance across indoor and outdoor benchmarks with varying action spaces, highlighting the advantage of long-term planning in active mapping. 
Project page: 
{\hyphenchar\font=-1 \url{https://shiyao-li.github.io/magician/}}
\end{abstract}    
\vspace{-1cm}
\section{Introduction}
\label{sec:intro}

% \shiyaormk{I feel like we may have overlooked an important point: our occupancy model is pretrained and thus carries strong data priors about general 3D structures. Combined with the fast feed-forward rendering of Gaussians, our approach effectively integrates the strengths of both learning-based and non-learning paradigms. It benefits from the prior knowledge encoded in the pretrained model while retaining the efficiency of Gaussian rendering, without requiring additional training on newly captured images or depth maps during exploration.}

%Maybe we should add some sentences like "we study the task of active 3D mapping in unknown environments: planning a sequence of camera viewpoints that efficiently reconstruct the full 3D structure of a scene. Unlike SLAM, which focuses on "Where am I?" and "What do I see?", active mapping addresses "Where should I look next to best understand the environment?", assuming the agent’s pose is known."

% \vincentrmk{we should mention rgb-d in the intro -- or cant you do as in macarons and use stereo reconstructions?  Do the concurrent methods use rgb-d cameras?}

Active mapping is a long-standing problem in computer vision and robotics~\cite{yamauchi1997frontier}, addressing the critical question: ``How should a mobile agent move to best reconstruct an unknown environment?''
Unlike SLAM~\cite{mur2015orb, davison2007monoslam} which focuses on camera localization and passive reconstruction, active mapping typically assumes known camera poses and emphasizes optimal viewpoint selection to enable efficient 3D reconstruction of complex scenes, minimizing exploration time while maximizing map quality.

To select the next best viewpoint, a variety of criteria have been proposed, such as information gain~\cite{bai2016information, bourgault2002information, stachniss2005information,isler2016information}, the Fisher information~\cite{jiang2024fisherrf}, or volumetric uncertainty~\cite{lee2022uncertainty}.
Among these, the surface coverage gain~\cite{mendoza2020nbvnet, zeng2020pcnbv, guedon2022scone, guedon2023macarons} has emerged as a state-of-the-art criterion due to its advantage in explicitly guiding the agent towards exhaustive exploration of the environment.

However, most existing active mapping methods only locally optimize the chosen criterion by iteratively predicting only the next best single pose~\cite{guedon2023macarons, chen2025gleam} or a short series of poses~\cite{li2025nextbestpath}.
Such greedy, short-sighted approaches lead to suboptimal exploration and mapping, with the agent losing time in dead-ends or performing unnecessary back-and-forth motions, as evidenced in prior literature~\cite{li2025nextbestpath} and confirmed by our experiments. 
% \shizhermk{Do we plan to show the trajectory comparisons? Good to refer the figure here.} \shiyaormk{we will show this in the supplemantary}

\shiyao{
} 

It is therefore essential to move beyond local pose optimization and employ \emph{long-term planning} to find globally efficient trajectories that cover more of the scene in less time.
In other words, optimizing the total accumulated surface coverage gain over a long trajectory rather than a single next pose.
Nevertheless, planning long-horizon trajectories is profoundly challenging.
First, the problem inherently suffers from the combinatorial explosion of possible trajectories, even  with known scene geometry~\cite{roberts2017submodular}.
Second, the required surface coverage gain for unknown future poses must be computed in an environment that is not yet fully observed.
Third, traditional methods~\cite{guedon2022scone,guedon2023macarons} for estimating this gain are inefficient for quick evaluation of numerous candidate viewpoints.
This leads to the central, chicken-and-egg question: \emph{How can we efficiently plan an optimal, long-term trajectory to map a scene when the knowledge required for planning (\ie, the map itself) is not yet known?}

Inspired by the human capability to rapidly infer the structure of unfamiliar environments by imagining unseen regions and planning exploration accordingly, we address this chicken-and-egg problem by introducing ``I\textbf{ma}gined \textbf{G}auss\textbf{i}ans'' for a\textbf{c}t\textbf{i}ve m\textbf{a}ppi\textbf{n}g (\textbf{\ourmodel{}}).
Our approach leverages a pre-trained volume occupancy network~\cite{guedon2023macarons} which encodes strong structural priors and predicts the probabilistic occupancy field for both seen and unseen regions based on past observations.
While this field allows us to infer unseen areas, serving as a world model for planning, its direct volumetric integration is computationally expensive, which makes it infeasible for efficient long-term planning.
To mitigate this cost, we propose \textbf{Imagined Gaussians} - a 3D scene representation generated by sampling the 3D space based on this occupancy network.
By using the predicted probability of each Gaussian to be occupied as opacity, we establish that the new surface coverage gain can be efficiently estimated by rendering these Imagined Gaussians from any candidate camera pose.
% This rendering procedure is faster by several orders of magnitude than the prior method~\cite{guedon2023macarons}, which relies on a second network and Monte Carlo sampling.

% \vincentrmk{we should give the times (or an estimation of them) somewhere in the paper} \shiyaormk{i will add this in the method sec}

This radical speedup in gain computation for any poses allows us to finally plan a long-term trajectory: we maximize the accumulated surface coverage by performing a tree search of the possible future moves. 
This integration makes the tree search highly efficient and tractable, despite the inherent combinatorial complexity of long-term planning.
We regularly update our Imagined Gaussians with new observations and re-run the tree search to refine the planned trajectory in a closed loop.
Our approach \ourmodel{} outperforms the state-of-the-art methods~\cite{yan2023active, feng2024naruto, chang2017matterport3d, li2025nextbestpath} on both outdoor and indoor benchmarks, \eg, achieving over 10\% scene coverage improvement on the challenging Macarons++ benchmark.

% Moreover, by contrast with many active mapping methods which are restricted to planar 2D trajectories, our approach scales to unconstrained 3D trajectories, allowing us to map unknown complex 3D environments. \shizhermk{Is this true? I thought previous methods like MACARONS also predict 3d trajectories.} \shiyaormk{many existing methods are focusing on indoor which is 2d, but here i guess we can say: We are the first to achieve a unified solution for active mapping in indoor and outdoor settings.}

In summary, our contributions are as follows:
\begin{itemize}

\item We introduce the first framework \ourmodel{}, to our knowledge, capable of generating long-horizon trajectories for active 3D mapping, which addresses inherent limitations of greedy, short-term viewpoint selection.

\item We propose Imagined Gaussians derived from a neural occupancy field to enable efficient and reliable coverage gain prediction from new viewpoints in unknown scenes, supporting feasible long-term planning with tree search.

\item \ourmodel{} attains state-of-the-art performance in both indoor and outdoor environments, showcasing robust adaptability to diverse action spaces.
\end{itemize}

\section{Related work}
\label{sec:related_work}

\begin{figure*}[t]
\centering
\includegraphics[width=0.99\textwidth]{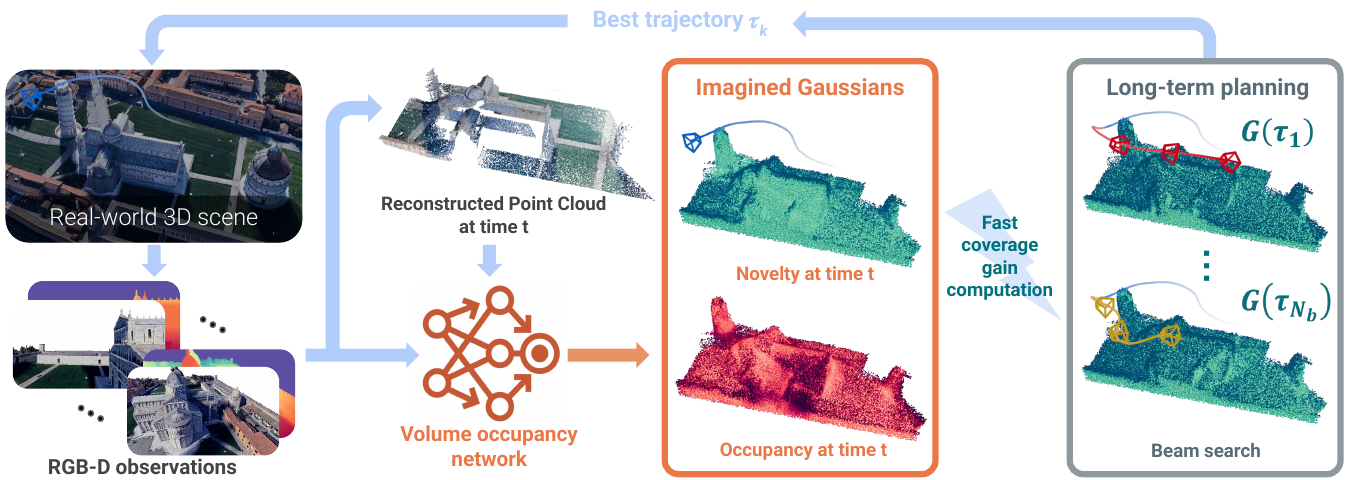}
\vspace{-1mm}
\caption{\textbf{Overview of the proposed \ourmodel{} framework.} 
At time $t$, we first predict the occupancy field using the occupancy model and update the Imagined Gaussians.
We can then efficiently estimate the coverage gain and apply beam search to plan $N_b$ candidate trajectories, selecting the one with the highest expected gain. The agent then executes the first $N_f$ actions of the best trajectory $\tau_k$ of length $N_d$ before repeating this process in the next planning loop.
In this figure, lighter colors in the Imagined Gaussians indicate higher novelty, while darker colors correspond to previously observed areas.
The first trajectory darkens the novelty field the most, representing the optimal path at time $t$.}
\label{fig:pipeline}
\end{figure*}

% Active mapping lies at the intersection of autonomous exploration and 3D reconstruction. The goal is to enable a mobile robot to efficiently explore and reconstruct an unknown environment within a limited time or motion budget. 

\noindent\textbf{Trajectory Planning in Active Mapping.}
Early approaches mainly relied on carefully designed heuristic criteria~\cite{nextscan, 20next, bourgault2002information, bai2016information, stachniss2005information} to guide exploration, such as selecting the next-best-view (NBV)~\cite{banta2000next, nbvp, 20next} or frontiers~\cite{yamauchi1997frontier, fastfrontier, heng2015efficient, batinovic2021multi}, or combining both strategies~\cite{cao2021tare, cao2020hierarchical}. However, these methods heavily depend on accurate environment modeling and handcrafted scoring functions.
Recent works~\cite{zeng2020pcnbv, guedon2023macarons, guedon2022scone} improve NBV selection via learning-based prediction of coverage gain, greedily choosing the most informative views. Yet, these myopic strategies still struggle with global coverage in complex environments or flexible action space due to the lack of long-term planning.
Beyond local NBV selection, some methods~\cite{feng2024naruto, jin2025activegs, chen2025activegamer, li2025activesplat} score candidate viewpoints and use classical planners (e.g., RRT~\cite{rrt}, A*~\cite{astar}) to reach them, but this decoupled design overlooks reconstruction gains accumulated along the path, leading to inefficiency under limited travel or time budgets.
Notably, only a few studies explore trajectory-level optimization in active mapping. For instance, FisherRF~\cite{jiang2024fisherrf} generates paths to multiple frontier targets and selects the most informative one, but still relies on frontiers. NextBestPath~\cite{li2025nextbestpath} learn to predict coverage gain along the shortest path between two viewpoints, though its performance remains sensitive to the quality and diversity of training data, limiting generalization.
In contrast, our method efficiently estimates coverage gain and performs tree-based long-term planning to find the optimal trajectory under a limited motion budget, achieving superior coverage efficiency.

% \todo{
% Explain that we use the occupancy module of MACARONS because it's not a voxel-based model (and we want to avoid voxels), it's not so easy to find an occupancy networl like that, etc.
% }

\noindent\textbf{Scene Representation in Active Mapping.}
Modeling the environment is crucial for effective active mapping. Traditional point cloud or voxel representations~\cite{zhou2021fuel, nbvp} are costly and resolution-limited, while image-based projections~\cite{li2025nextbestpath, chen2025gleam} simplify learning but remain confined to indoor scenes and lack full-coverage guarantees.
Building on advances in NeRF~\cite{mildenhall2020nerf} and 3D Gaussian Splatting (3D GS)~\cite{kerbl3Dgaussians, guedon2024sugar, 2dgs}, recent works have explored radiance-based representations for active mapping.
NeRF-based methods exploit internal training cues such as loss gradients or uncertainty~\cite{yan2023active, feng2024naruto, pan2022activenerf} to guide view selection, while GS greatly improves rendering efficiency, enabling compact differentiable scene representations.
Several GS-based methods~\cite{jiang2024fisherrf, jin2025activegs, chen2025activegamer, li2025activesplat} evaluate candidate views via information gain, confidence, or rendered density, but all require frequent Gaussian updates during exploration.
However, all these methods require frequent updates to the Gaussian representation during exploration.
Unlike these approaches, our method predicts a 3D occupancy proxy field and converts it into Imagined Gaussians, leveraging 3D Gaussians’ fast feed-forward rendering efficiency while avoiding costly gradient-based updates.

\section{Method}
\label{sec:surface-coverage-as-a-volumetric-integral}

\subsection{Problem Definition}

Active 3D mapping aims to explore an unknown environment using a mobile agent (\eg, a drone or ground robot) to achieve a high-fidelity 3D reconstruction in the minimum possible time or shortest trajectory length.
Starting from an arbitrary initial pose, the agent operates in an iterative perception-action loop.
At each time step $t$, it acquires an RGB-D observation $I_t$ from its camera pose $\cam_t$. 
Based on the current understanding of the environment, the agent must then actively select the next viewpoint $\cam_{t+1} \in \mathrm{SE}(3)$ in its vicinity, which defines its subsequent 3D position and orientation.
The agent continues the loop until reaching a maximum time $T$.

\begin{figure}
\includegraphics[width=\linewidth]{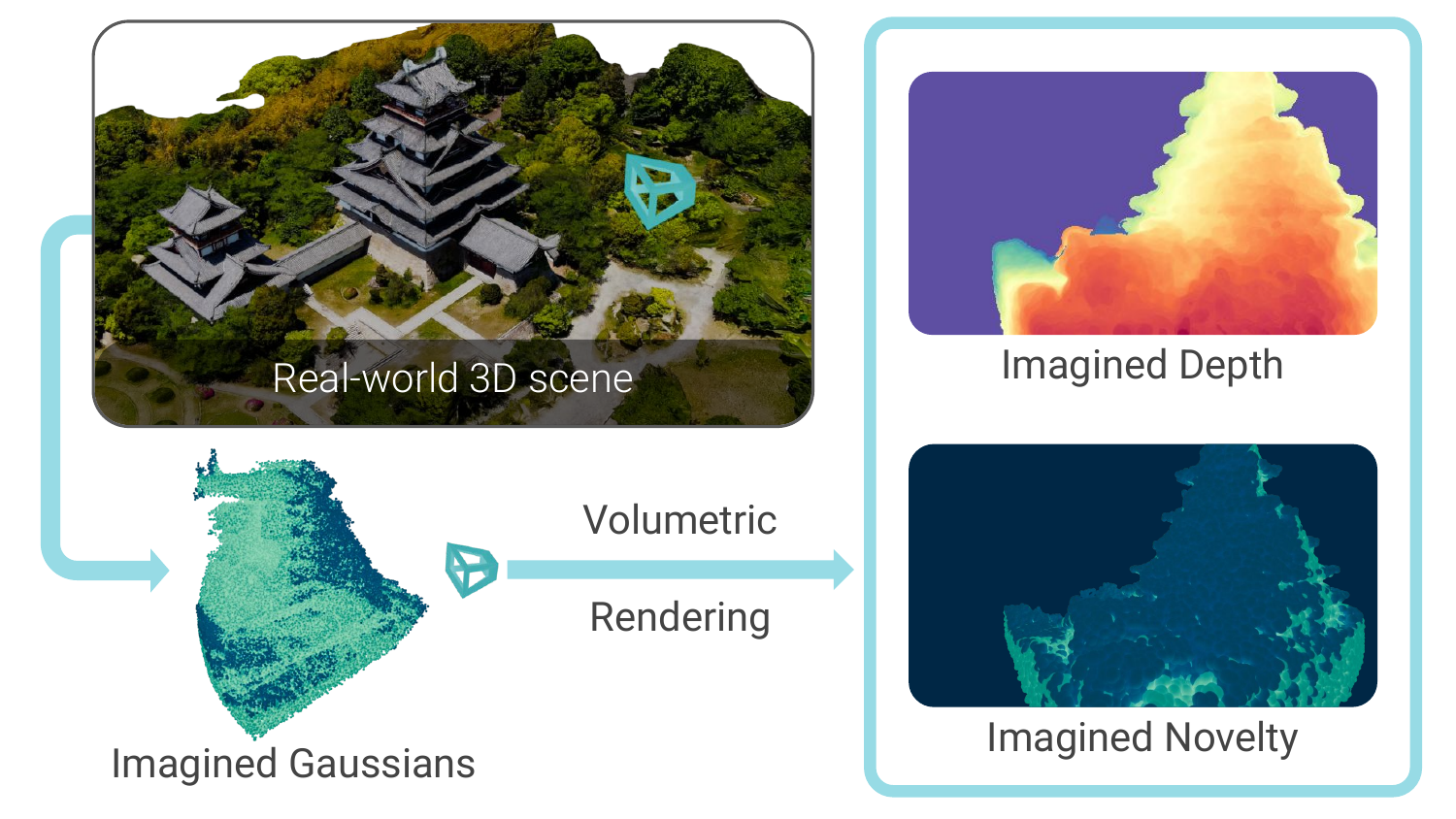}
\vspace{-0.5cm}
\caption{\textbf{Computing coverage gain with Imagined Gaussians.} During beam search, we evaluate candidate poses by rendering novelty maps from the Imagined Gaussians to compute the coverage gain. The corresponding depth maps are then used to update the novelty $\nov$ of Gaussians within a depth tolerance $\epsilon_d$.
% \shizhermk{1. put this figure before gaussian evolution figure. 2. improve the caption, it is not related to beam search. the figure is referred in sec 3.4.}
}
\label{fig:imagined_gaussians}
\end{figure}
\subsection{Optimizing Long-term Surface Coverage Gain}

The surface coverage gain~\citep{mendoza2020nbvnet, zeng2020pcnbv, guedon2022scone, guedon2023macarons} is one of the state-of-the-art criteria in active mapping for selecting the next best viewpoint.
% \vincentrmk{i am not sure about 'widely adopted', this could shock the authors of fisherrf if they review our paper-the sentence in the introduction was nicer}
It quantifies the amount of new, unobserved surface area revealed from a candidate camera pose $\cam$ relative to the previously visited poses $\bC_t = \{\cam_1, \ldots, \cam_t\}$:
\begin{equation}
G(\cam) = 
\int_{\partial \calE \cap \VF(\cam)} 
\sigma(\vecx) \cdot
o(\vecx, \cam) \cdot
\gamma(\vecx | \bC_t)
d\vecx \> ,
\label{eq:surface-coverage-gain}
\end{equation}
where $\calE\subset\IR^3$ is the occupied 3D space. The surface integral is performed over the intersection of the true scene surface $\partial \calE$ and the camera view frustum $\VF(\cam)$. 
Specifically, $\sigma(\vecx) = \mathbf{1}_{\calE}(\vecx)$ indicates whether point $\vecx$ is occupied;
$o(\vecx, \cam)$ equals 0 if point $\vecx$ is occluded from camera $\cam_t$ and 1 otherwise;
and $\gamma(\vecx | \bC_t) \in \{0, 1\}$, referred to as the \emph{novelty} indicator, equals 1 if and only if the point $\vecx$ has not been previously observed in $\bC_t$.

Prior approaches~\citep{guedon2022scone,guedon2023macarons} using surface coverage gain are often short-sighted, greedily optimizing the immediate gain $G(\cam_{t+1})$.
This leads to locally optimal but globally inefficient exploration paths.
To address this limitation, we propose an exploration objective to optimize the \textbf{total accumulated surface coverage gain} $G(\tau)=\sum_{i=1}^{N_d} G(\cam_{t+i})$ over a long-term trajectory $\tau = \{\cam_{t+1}, \cdots, \cam_{t+N_d}\}$ of length $N_d$.

Solving this long-term optimization problem presents three primary challenges:
i) Direct computation of the ideal $G(\cam)$ is intractable because the true scene surface $\partial \mathcal{E}$ and its occupancy $\sigma(\vecx)$ are unknown during exploration. 
ii) Due to the high-dimensional pose space, a highly efficient method is required to measure $G(\cam)$ for numerous candidate viewpoints.
iii) We need a scalable planning approach to generate an optimal, long-horizon trajectory $\tau$ that maximizes the accumulated gain $\mathcal{G}(\tau)$ without exhaustive searching.

To address these challenges, we introduce \ourmodel{} as illustrated in~\Cref{fig:pipeline}.
First, we employ a pre-trained neural occupancy model to estimate the geometry in both seen and unseen areas  (\Cref{sec:method_occupancy}). 
Second, we propose Imagined Gaussians, which use volumetric rendering to measure $G(\cam)$ with high efficiency (\Cref{sec:method:imagined_gaussian}). 
Finally, we integrate this rapid gain calculation into an efficient tree-search method to enable robust long-term trajectory planning (\Cref{sec:method_planning}).

\begin{figure*}[t]
  \centering
  \begin{subfigure}{0.33\linewidth}
    \centering
    % left bottom right top
    \includegraphics[width=\linewidth, trim= 4cm 6cm 4.5cm 0cm,clip]{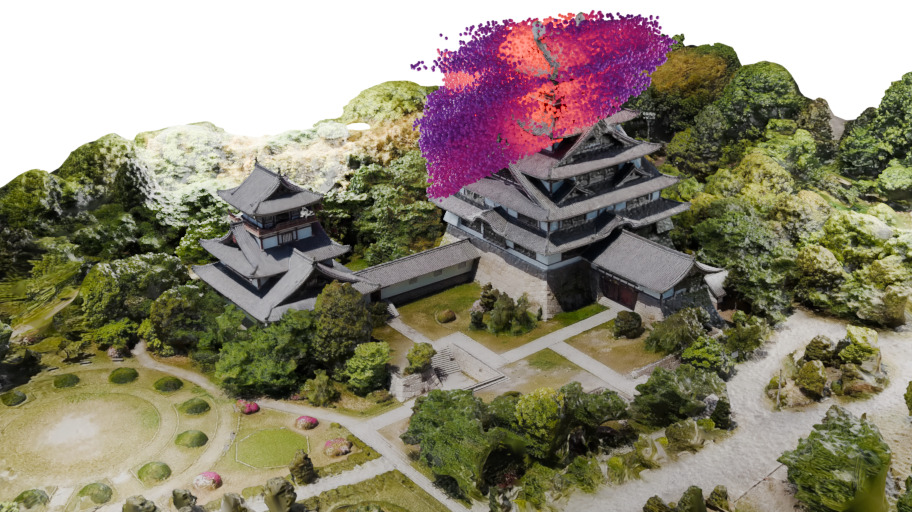}
    \caption{Imagined Gaussians at $t_0$}
    \label{fig:sub-a}
  \end{subfigure}\hfill
  \begin{subfigure}{0.33\linewidth}
    \centering
    \includegraphics[width=\linewidth, trim= 4cm 6cm 4.5cm 0cm,clip]{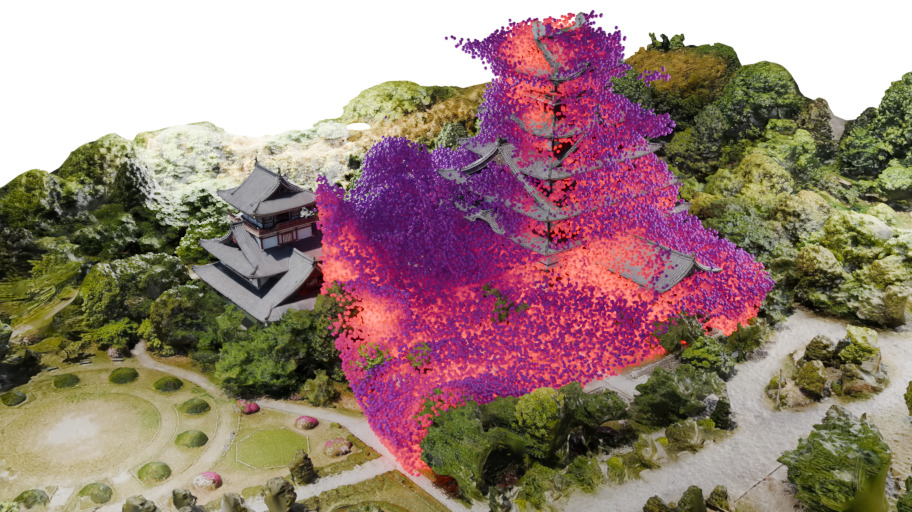}
    \caption{Imagined Gaussians at $t_1 > t_0$}
    \label{fig:sub-b}
  \end{subfigure}\hfill
  \begin{subfigure}{0.33\linewidth}
    \centering
    \includegraphics[width=\linewidth, trim= 4cm 6cm 4.5cm 0cm,clip]{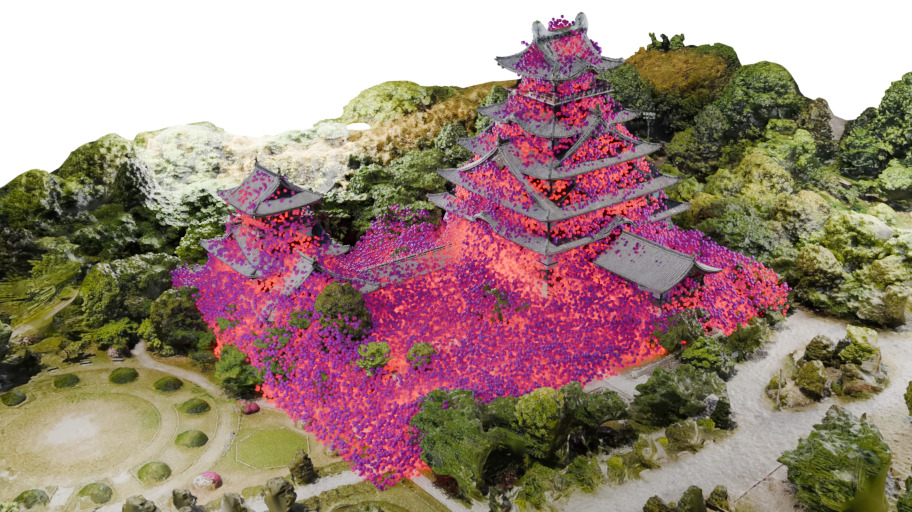}
    \caption{Imagined Gaussians at $t_2 > t_1$}
    \label{fig:sub-c}
  \end{subfigure}

  \caption{\textbf{Evolution of Imagined Gaussians Compared with Ground Truth Mesh.} The brighter the Gaussians, the higher their predicted occupancy. As exploration progresses (from left to right), our Imagined Gaussians increasingly align with the ground truth mesh, demonstrating improved environmental modeling.
  % \shizhermk{I prefer to mention this figure in the result section (ablation), not in the method section.}
  } 
  % \shiyaormk{but this figure is very fancy, will attract reviewers to read more}
  \label{fig:occ_field}
\end{figure*}

 % \antoine{The brighter the Gaussians, the higher their predicted occupancy.}\shiyaormk{maybe opacity? since here we already convert the occupancy points to gaussians}\antoinermk{We explain sooner in the paper that we use the occupancy as the opacity. I'm afraid opacity will be misleading as it refers to usual RGB Gaussians + since we spend a lot of time talking about occupancy prediction and network, I think it makes more sense to maintain the wording occupancy here to make it clear that these values are predicted by the network.} 
\subsection{Neural Occupancy Prediction}
\label{sec:method_occupancy}

We train a neural occupancy prediction model $\occ(\vecx | \bC_t)$ to estimate the true occupancy $\sigma(\vecx)$ in partially observed environments. Our model follows the architecture of prior work~\cite{guedon2022scone,guedon2023macarons}, which is a multi-layer transformer.
The network takes as input the point $\vecx$, the reconstructed surface point cloud and previous poses.
It outputs a probability field where $\occ(\vecx | \bC_t) \in [0, 1]$ represents the likelihood that a point $\vecx$ is occupied. 
The occupancy model is first pre-trained on ShapeNet~\cite{chang2015shapenet} and then fine-tuned on 3D scenes~\cite{guedon2023macarons}, and thus carries strong prior knowledge about general 3D structures. This occupancy model also allows us to plan collision-free trajectories.

It is important to note that our approach is generalizable and can incorporate any occupancy network.

% To address these challenges, we propose using volumetric rendering to enable orders of magnitude faster computation of $G(\cam)$ without sacrificing accuracy.
% Moreover, we propose to generate imagined Gaussians to compute the coverage gain more effectively, and then applly tree-search to enable long-term planning.

\subsection{Imagined Gaussians}
\label{sec:method:imagined_gaussian}
With the probabilistic occupancy field $\occ(\vecx | \bC_t)$ estimated by our model, we next describe how to efficiently compute the coverage gain $G(\cam)$ for a candidate viewpoint.  

Prior work~\cite{guedon2022scone, guedon2023macarons} approximates Eq.~\eqref{eq:surface-coverage-gain} with a volumetric Monte Carlo integral over the camera frustum $\VF(\cam)$:
\begin{equation}
G(\cam) \approx \int_{\VF(\cam)} 
\occ(\vecx | \bC_t) \cdot 
\occl(\vecx, \cam) \cdot
\nov(\vecx | \bC_t) \cdot 
d\vecx \> ,
\label{eq:volumetric-coverage-integration}
\end{equation}
where the product $\occl(\vecx, \cam)\nov(\vecx | \bC_t)$ is approximated by a second neural network. However, computing this integral via Monte Carlo sampling requires repeatedly querying both networks on dense 3D points, making it computationally prohibitive for long-term exploration.

\noindent\textbf{Volumetric rendering for coverage gain estimation.}
Our key insight is that Eq.~\eqref{eq:volumetric-coverage-integration} shares the same structure as the volumetric rendering equation~\cite{mildenhall2020nerf} used in NeRF and 3D Gaussian Splatting:
\begin{equation}
I(\pix) = \int_{0}^{+\infty} q(\veco+s\vecd) \cdot T(s; \veco, \vecd) \cdot f(\veco+s\vecd, \vecd) ds
\label{eq:volumetric-rendering-equation}
\end{equation}
where $q$, $T$, and $f$ represent density, transmittance, and color along a ray $\{\veco+s\vecd\}$ passing through pixel $\pix$. Transmittance $T$ equals 1 in empty space and quickly decays to 0 after reaching the first occupied space. It can thus be interpreted as a relaxed version of the occlusion function $\occl(\bx,\bc)$ in Eq.~\eqref{eq:volumetric-coverage-integration}. Similarly, density field $q(\vecx)$ describes the opacity of the scene and can be used to represent the probabilistic occupancy field $\occ(\vecx)$ of Eq.~\eqref{eq:volumetric-coverage-integration}. Vector field $f(\vecx)$ typically describes the RGB color emitted by point $\vecx$, but we use it here to represent novelty $\nov(\vecx | \bC_t)$. 

By correspondence, $q \leftrightarrow \occ$, $T \leftrightarrow \occl$, $f \leftrightarrow \nov$, and Eq.~\eqref{eq:volumetric-rendering-equation} becomes:
\begin{equation}
I(\pix) = \int_{0}^{+\infty} 
\!\!\!\!\!\!
\occ(\veco+s\vecd | \bC_t) \cdot 
\occl(\veco+s\vecd, \cam) \cdot 
\nov(\veco+s\vecd | \bC_t) 
ds \> ,
\label{eq:volumetric-rendering-equation-with-coverage}
\end{equation}
allowing to estimate with volumetric rendering the coverage gain over an infinitesimal surface patch corresponding to pixel $\pix$.
Summing over all pixels yields the full coverage gain $G(\cam)$ over $\VF(\cam)$.
This formulation eliminates Monte Carlo sampling, leverages GPU-accelerated volumetric rendering, and requires only a single occupancy network, leading to orders-of-magnitude faster computation.

\noindent\textbf{Imagined Gaussians for volumetric rendering.}
To instantiate Eq.~\eqref{eq:volumetric-rendering-equation-with-coverage}, we represent the scene as a collection of 3D Gaussian primitives centered on \textit{proxy points} from the occupancy network $\occ(\vecx | \bC_t)$ of~\cite{guedon2023macarons}. These proxy points are randomly sampled with higher density inside of the exploration bounding box. We use isotropic Gaussians with radius equal to half the distance to the nearest neighbor. Following Eq.~\eqref{eq:volumetric-rendering-equation-with-coverage}, Gaussian opacities encode occupancy probabilities $\occ(\vecx | \bC_t)$ and colors encode binary novelty $\nov \in \{0, 1\}$. A Gaussian is marked observed if its center's distance from a previous pose $\cam$ matches the rendered depth within tolerance $\epsilon_d$, as illustrated in \Cref{fig:imagined_gaussians}.

We call this scalable volumetric representation \emph{Imagined Gaussians}, as some Gaussians have not been observed yet and their occupancies are only predicted. It supports fast rasterisation and accurate coverage computation, serving as a foundation for long-term planning.

\noindent\textbf{Fast coverage gain computation with Imagined Gaussians.}
For any candidate pose, we compute its coverage gain by first rendering a novelty map from the current Imagined Gaussian state via volumetric rendering (Eq.~\eqref{eq:volumetric-rendering-equation-with-coverage}), then summing the rendered novelty over all pixels. The coverage gain is computed only over valid regions where depth is available, ensuring we evaluate observable surfaces.

\subsection{Long-Term Planning}
\label{sec:method_planning}

Our fast coverage gain computation, enabled by Imagined Gaussians, directly facilitates efficient long-term trajectory planning. Planning such trajectories is non-trivial: the agent must anticipate future observations along the path to avoid redundant views and identify the globally  efficient exploration paths. To address this, we employ a beam search strategy that incrementally expands the exploration trajectory over candidate camera poses.

We periodically execute the beam search over the next $N_d$ possible moves to find the optimal continuation trajectory $\tau = \{\cam_{t+1}, \ldots, \cam_{t+N_d}\}$. 
Assume we have $N_b$ beams. Each beam represents a possible future trajectory and maintains its own independent copy of the Imagined Gaussian state. 
At each iteration, we expand every active beam by one move. This expansion involves enumerating all camera poses reachable from the current trajectory endpoint using the available agent actions (e.g., translation and rotation primitives).
We calculate the coverage gain with the corresponding Imagined Gaussians for each move and only keep the top $N_b$ beams for the next expansion.

\begin{figure*}[!ht]

\setlength{\fboxsep}{0pt}   % padding
\setlength{\fboxrule}{1.pt}% border thickness

\vspace*{7mm}

\centering

% Neushchwanstein
\begin{minipage}[b]{0.196\textwidth}
    \centering
    \includegraphics[width=\linewidth]{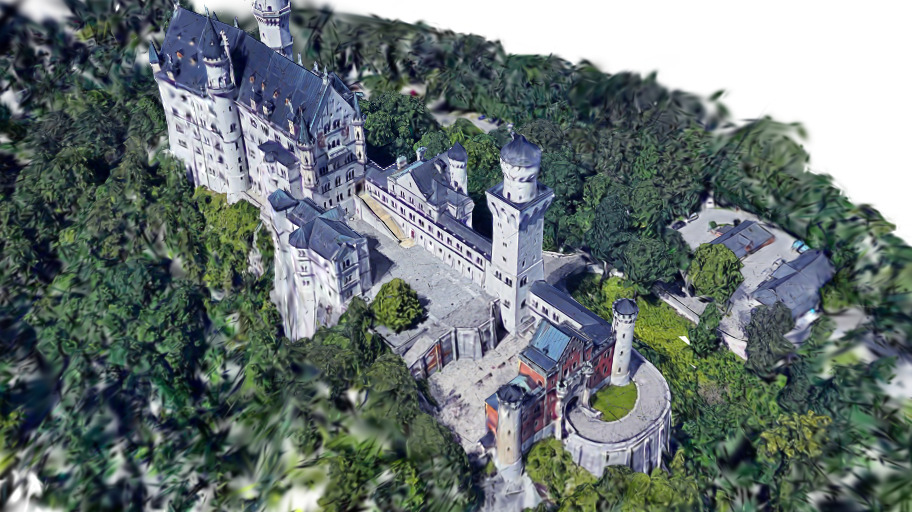}
    \includegraphics[width=\linewidth]{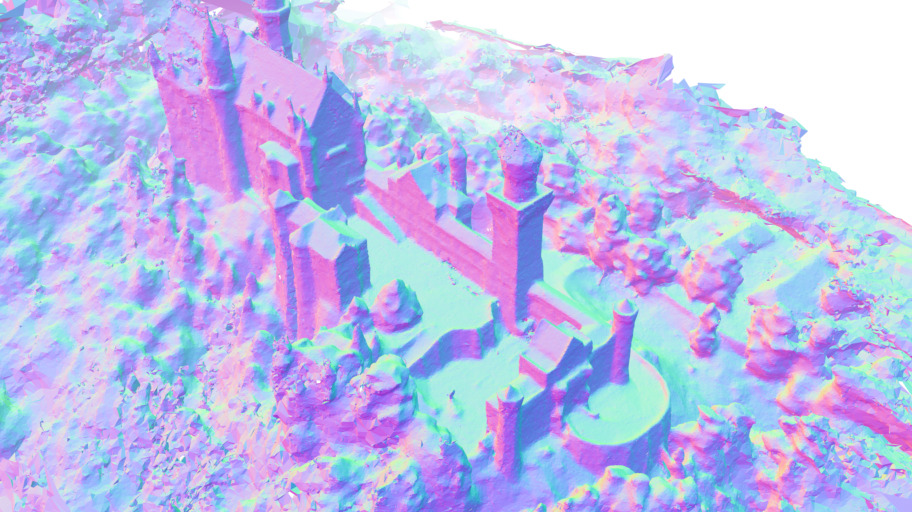}
    \vfill
    \scriptsize (a) Neuschwanstein Castle
\end{minipage}
\hfill
% Colosseum
\begin{minipage}[b]{0.196\textwidth}
    \centering
    \includegraphics[width=\linewidth]{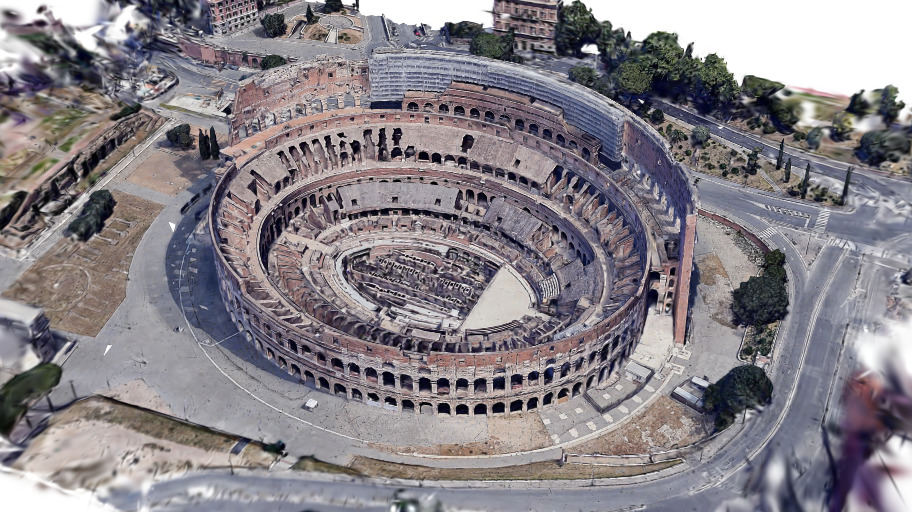}
    \includegraphics[width=\linewidth]{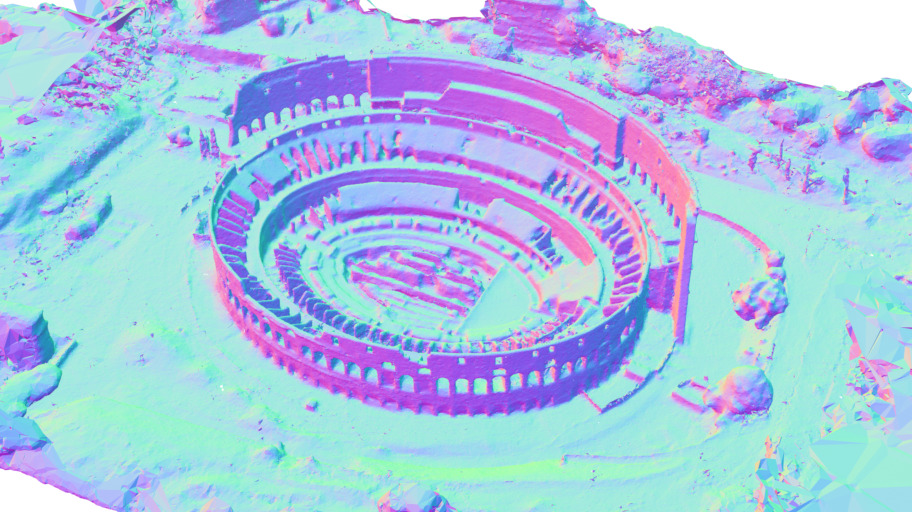}
    \vfill
    \scriptsize (b) Colosseum
\end{minipage}
\hfill
% Fushimi
\begin{minipage}[b]{0.196\textwidth}
    \centering
    \includegraphics[width=\linewidth]{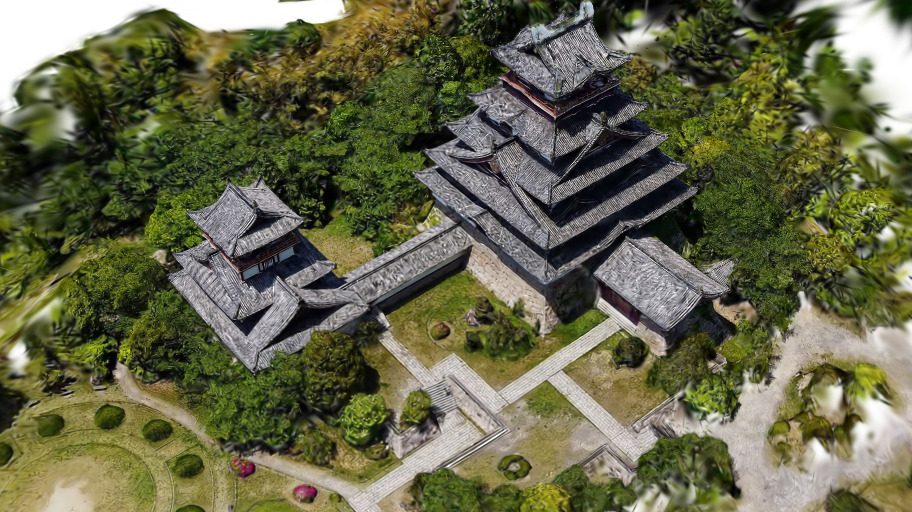}
    \includegraphics[width=\linewidth]{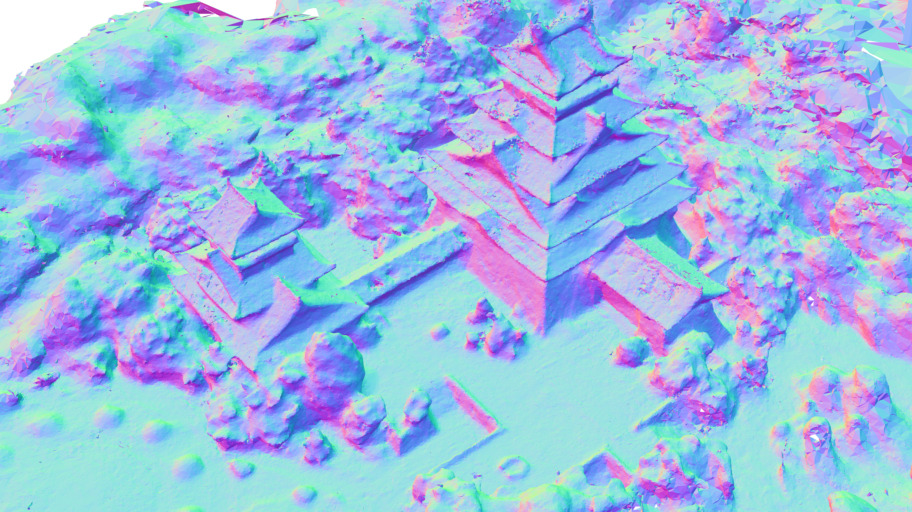}
    \vfill
    \scriptsize (c) Fushimi Castle
\end{minipage}
\hfill
% Alhambra
\begin{minipage}[b]{0.196\textwidth}
    \centering
    \includegraphics[width=\linewidth]{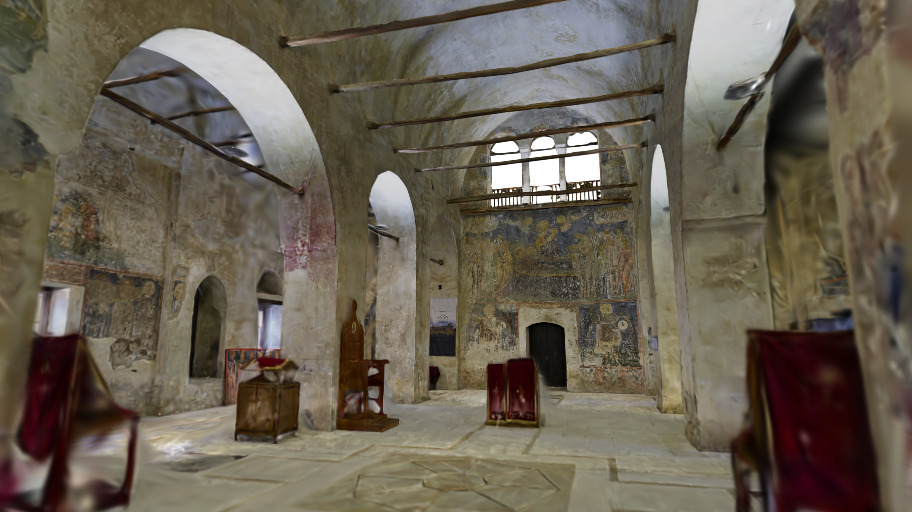}
    \includegraphics[width=\linewidth]{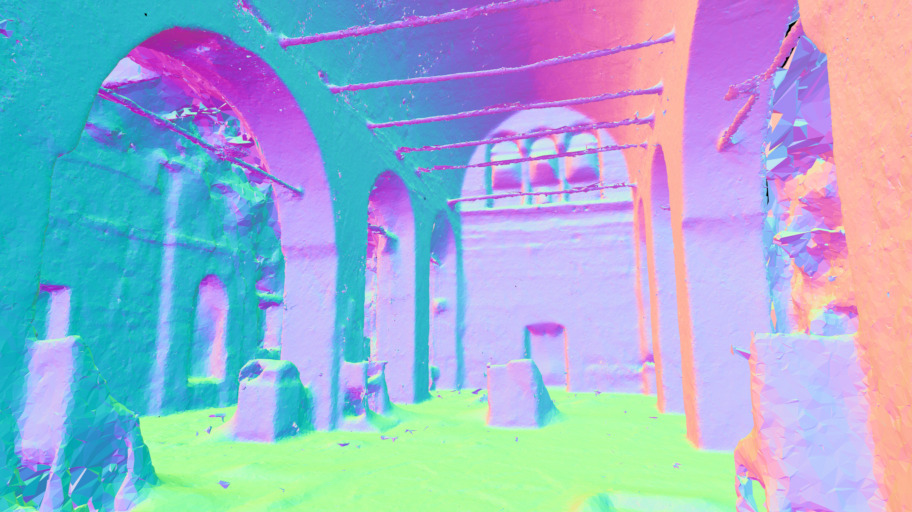}
    \vfill
    \scriptsize (d) St. Sofia Church
\end{minipage}
% Church
\hfill
\begin{minipage}[b]{0.196\textwidth}
    \centering
    \includegraphics[width=\linewidth]{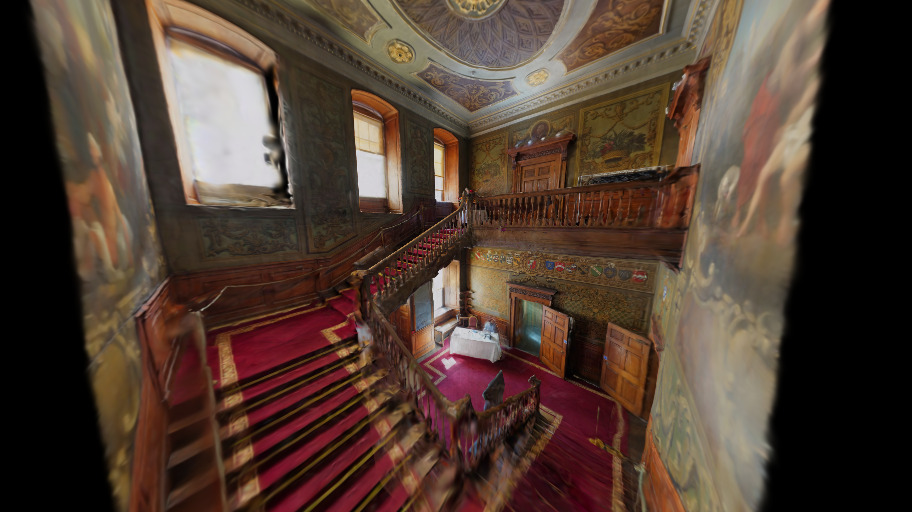}
    \includegraphics[width=\linewidth]{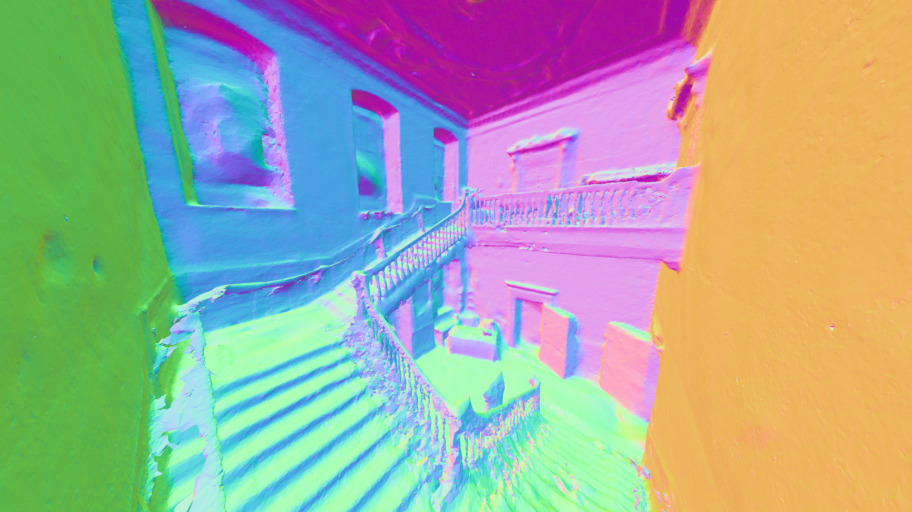}
    \vfill
    \scriptsize (e) Barts
\end{minipage}

\caption{
%\vincent{
\textbf{3D reconstructions obtained with our trajectories.} 
%
% \todo{Add a first row with trajectories?}
%
We show Gaussian splatting renderings (top row) and normal maps of the reconstructed meshes (bottom row) after applying Mesh-In-the-Loop Gaussian Splatting~\cite{guedon2025milo} on 100 RGB images collected along our trajectories.
The trajectories output by our method cover the entire scene surfaces, resulting in complete and accurate surface meshes. 
% \vincentrmk{is it possible to replace one of the outdoor scenes (maybe Alhambra) by an indoor scene, to balance things out? Also, this figure should be moved at the end of the paper.}
% To evaluate the coverage of our camera trajectories, we optimize a 3D Gaussian splatting representation on 100 images collected along the trajectory.
% %
% Specifically, we apply Mesh-In-the-Loop Gaussian splatting~\cite{guedon2025milo} to allow for both novel view synthesis and surface mesh extraction.
% %
% The top row 
% % illustrates our camera trajectories, the middle row
% showcases renderings obtained with 3D Gaussians, and the bottom row presents normal maps of the extracted surface meshes.
% %
% %
% We believe our approach could facilitate the practical reconstruction of 3D environments from RGB images, as well as their deployment in 3D simulation softwares.
% }
}

\vspace*{0mm}
\label{fig:qualitative_results}
\end{figure*} 

During beam search, we keep the Gaussian parameters frozen except for their novelty values encoded by colors. When a Gaussian is observed from a candidate pose---as determined by rendering its depth and checking visibility---we update its novelty from 1 to 0 for that beam's state. This ensures that when rendering novelty maps for subsequent candidate poses in the trajectory, the contribution of that Gaussian is automatically reduced through the volumetric rendering equation, thereby excluding it from coverage gain computation for the remainder of that trajectory. Crucially, each beam maintains its own independent Gaussian state, allowing parallel exploration of different trajectory hypotheses with distinct observation histories.
The value of a trajectory is the sum of coverage gains $\sum_{i=1}^{N_d} G_{\text{rendered}}(\cam_i)$ along its $N_d$ steps. 

After beam search completes, we execute the first $N_f \leq N_d$ steps of the best trajectory, moving the agent and capturing new observations. 
We then update the Imagined Gaussians based on these real observations: observed Gaussians have their opacities refined by the occupancy network as illustrated in \Cref{fig:occ_field}, and their novelty values are set to 0.
We iteratively perform this perception-planning-action loop until reaching the maximum timesteps.

\section{Experiments}
\label{sec:experiments}

\subsection{Experiments Setup}

\begin{table}[t]
\centering
\footnotesize
\tabcolsep=0.08cm
\caption{Configuration details on MP3D and Macarons++ benchmarks for active mapping evaluation.}
\label{tab:dataset_setup}
\begin{tabular}{lccc} \toprule
 & \multicolumn{2}{c}{\textbf{MP3D}} & \textbf{Macarons++} \\
 & \textbf{Wheeled Robot} & \textbf{Drone} & \textbf{Drone} \\ \midrule
\textbf{Action space} & \begin{tabular}[c]{@{}l@{}}Forward (6.5 cm), \\ Turn ($\pm 10^\circ$)\end{tabular} & 6 DoF & 6 DoF \\
\textbf{Camera Height} & 1.25 m above floor & N/A (Free 3D) & N/A (Free 3D) \\
\textbf{Image Resolution} & $256 \times 256$ & $1200 \times 680$ & $456 \times 256$\\
\textbf{Field of View} & $90^\circ \times 90^\circ$ & $60^\circ \times 90^\circ$ & $60^\circ \times 91.6^\circ$ \\
\textbf{Total Steps} & $1,000$ or $2,000$ & $5,000$ & $100$\\
\textbf{Prior Work} & \cite{yan2023active, ramakrishnan2020occupancy, georgakis2022uncertainty, li2025nextbestpath} & \cite{chen2025activegamer, feng2024naruto} & \cite{guedon2022scone,guedon2023macarons}\\ \bottomrule
\end{tabular}
\end{table}

\textbf{Datasets.} 
We evaluate our method on two benchmarks: the Matterport3D (MP3D) dataset~\cite{chang2017matterport3d}, which contains indoor environments, and Macarons++, an extended version of the Macarons dataset~\cite{guedon2022scone, guedon2023macarons}.
Macarons++ includes large-scale 3D outdoor real-scan meshes in Macarons, and three new complex indoor scenes from Sketchfab, released under a Creative Commons license.

For MP3D, we follow prior work~\cite{yan2023active, li2025nextbestpath, chen2025activegamer, feng2024naruto} and use five scenes for evaluation. 
Since different studies adopt varying robot embodiments and action spaces, we ensure a fair comparison by evaluating our method under two commonly used configurations: a wheeled robot and a drone.
For the Macarons++ dataset, we follow the experimental setup used in MACARONS~\cite{guedon2023macarons}. Details are provided in~\Cref{tab:dataset_setup}.

\noindent
\textbf{Evaluation metrics.} Following \cite{guedon2023macarons, li2025nextbestpath, guedon2022scone}, we consider two metrics:
(1) \textbf{\textit{Final Coverage}}, which measures the overall scene coverage achieved at the end of the exploration trajectory; and
(2) \textbf{\textit{AUC}}, which evaluates the efficiency of the reconstruction process as the area under the curve of coverage over time. The surface coverage is computed using ground-truth meshes as in \cite{guedon2023macarons}. For each method, we evaluate five trajectories per scene using identical random initial camera poses to ensure fair comparison.

% \textbf{Evaluation metrics.} Following prior works~\cite{guedon2023macarons, li2025nextbestpath, guedon2022scone}, we use two key metrics to evaluate the performance of active mapping:
% (1) \textbf{\textit{Final Coverage}}, which measures the overall scene coverage achieved at the end of the exploration trajectory; and
% (2) \textbf{\textit{AUC}}, which evaluates the efficiency of the reconstruction process by computing the area under the curve of coverage over step. The surface coverage is computed using ground-truth meshes, consistent with prior works \cite{guedon2023macarons}. For each method, we evaluate five trajectories per scene using identical random initial camera poses to ensure fair comparison.

To further evaluate the quality of active mapping, we use 100 images collected from each trajectory and train 3D Gaussian representations for every method and scene. For evaluation, each scene is associated with a fixed set of novel-view images generated through submodular optimization, on which all methods are evaluated. Details are provided in the supplementary material. We perform rendering evaluation on these novel views and extract high-quality meshes from the trained 3D Gaussians using the state-of-the-art method MILo~\cite{guedon2025milo}. The reconstructed meshes are then compared with the ground-truth meshes to evaluate geometric accuracy, where the threshold for accuracy is set to 1\% of the diagonal length of each scene.

For comparison with prior studies on MP3D, we also consider the following metrics for scene coverage: (1) \textbf{\textit{Comp.(\%)}}, denoting the fraction of ground-truth vertices lying within 5 cm of any reconstructed observation, and (2) \textbf{\textit{Comp.(cm)}}, quantifying the average shortest distance from each ground-truth vertex to its nearest reconstructed point.

% To ensure fair comparison with prior studies on MP3D, we introduce an additional set of evaluation metrics for scene coverage: (1) \textbf{\textit{Comp.(\%)}}, denoting the fraction of ground-truth vertices lying within 5 cm of any reconstructed observation, and (2) \textbf{\textit{Comp.(cm)}}, quantifying the average shortest distance from each ground-truth vertex to its nearest reconstructed point.

\noindent
\textbf{Implementation details.} During exploration, we use the differentiable Gaussian rasterizer from RaDe-GS~\cite{zhang2024rade} to generate accurate depth maps with our imagined Gaussians. We set the beam width $N_b = 10$ and the planning horizon $N_d = 10$ steps, executing $N_f = 1$ step before replanning.

\subsection{Comparison with State-of-the-Art Methods}

\begin{figure*}[!ht]

\setlength{\fboxsep}{0pt}   % padding
\setlength{\fboxrule}{1.pt}% border thickness

\vspace*{7mm}

\centering

% FisherRF
\begin{minipage}[b]{0.33\textwidth}
    \centering
    {\includegraphics[width=\linewidth]{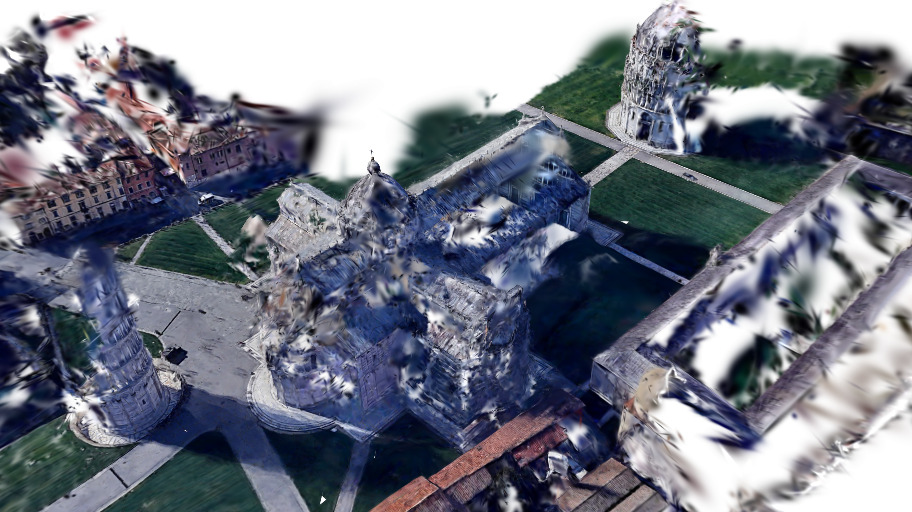}}
    {\includegraphics[width=\linewidth]{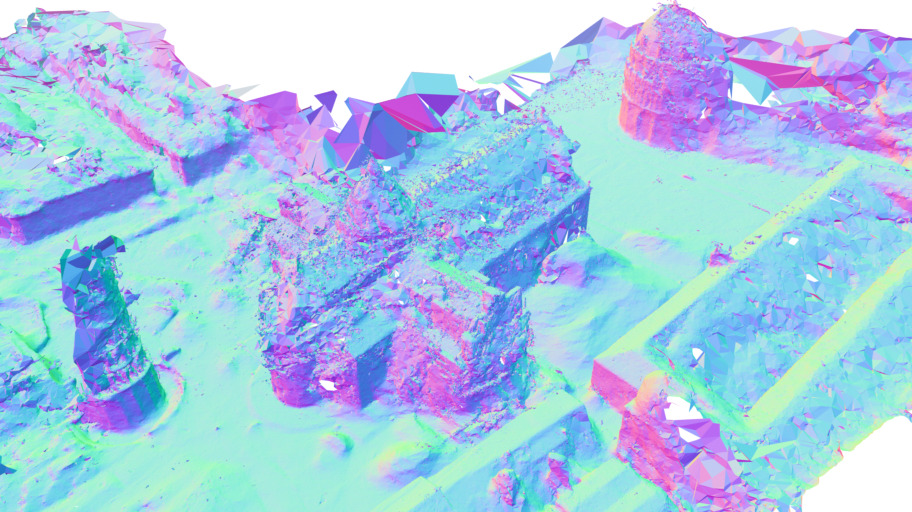}}
    % {\includegraphics[width=0.98\linewidth]{sec/images/render/stair_fisherf_cam1/rgb.jpg}}
    % {\includegraphics[width=0.98\linewidth]{sec/images/render/stair_fisherf_cam1/mesh_normal.jpg}}
    \vfill
    FisherRF~\cite{jiang2024fisherrf}
\end{minipage}
\hfill
% MACARONS
\begin{minipage}[b]{0.33\textwidth}
    \centering
    {\includegraphics[width=\linewidth]{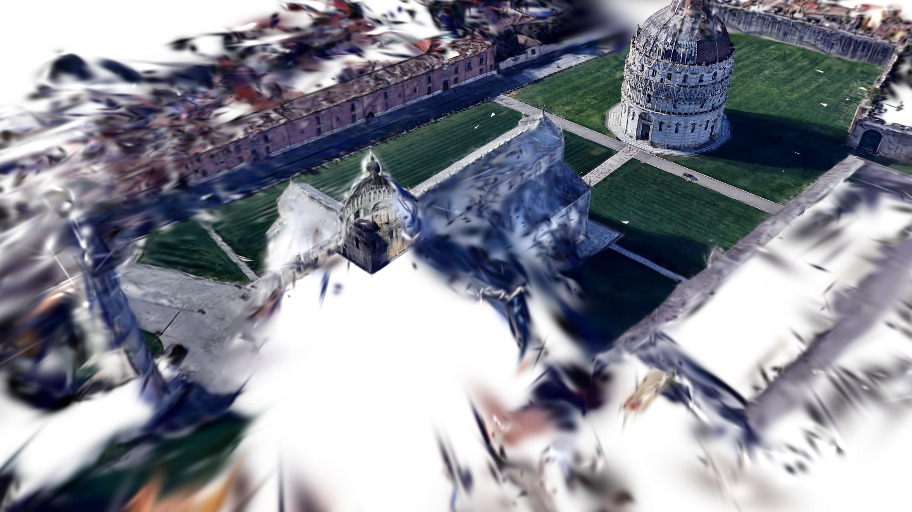}}
    {\includegraphics[width=\linewidth]{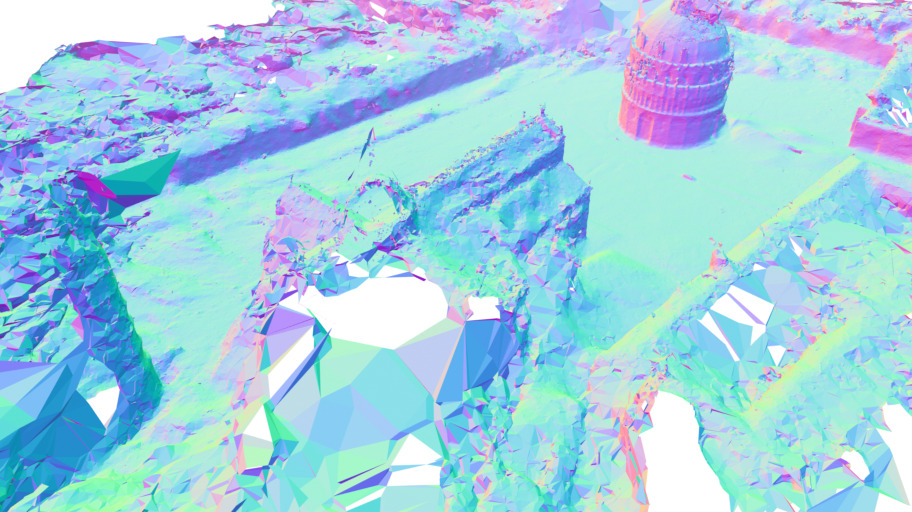}}
    % {\includegraphics[width=0.98\linewidth]{sec/images/render/stair_macarons_cam1/rgb.jpg}}
    % {\includegraphics[width=0.98\linewidth]{sec/images/render/stair_macarons_cam1/mesh_normal.jpg}}
    \vfill
    MACARONS~\cite{guedon2023macarons}
\end{minipage}
\hfill
% Ours
\begin{minipage}[b]{0.33\textwidth}
    \centering
    {\includegraphics[width=\linewidth]{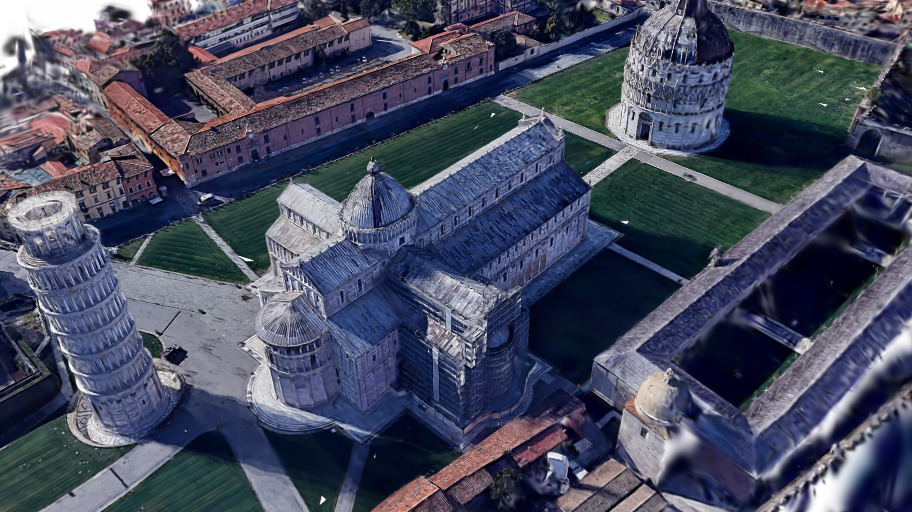}}
    {\includegraphics[width=\linewidth]{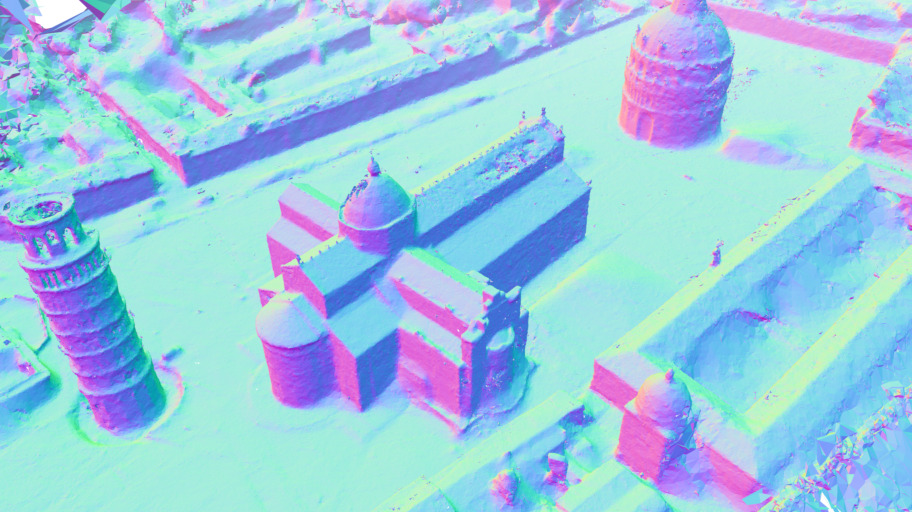}}
    % {\includegraphics[width=0.98\linewidth]{sec/images/render/stair_ours1_cam1/rgb.jpg}}
    % {\includegraphics[width=0.98\linewidth]{sec/images/render/stair_ours1_cam1/mesh_normal.jpg}}
    \vfill
    Ours
\end{minipage}

\caption{
% \vincent{
\textbf{Qualitative comparison of novel view synthesis (top row) and surface reconstruction (bottom row) in outdoor and indoor scenes.} 
For each method, 
% we show the reconstructed mesh and a normal map after Gaussian Splatting on 100 images collected along the trajectory.
we show RGB Gaussian splatting renderings and normal maps of reconstructed meshes after applying Mesh-In-the-Loop Gaussian Splatting~\cite{guedon2025milo} on 100 images collected along the trajectory.
The trajectories computed with our method produce more accurate and complete reconstructions, resulting in better rendering quality and preventing holes in reconstructed surfaces. 
% \shizhermk{better to put the figure nearby the corrresponding discussion in the text.}
% \shiyaormk{do we need to mention that we use reconstructed point cloud to initialize the gausians intro}
% \vincentrmk{the color schemes for the normals are not the same for the 2 examples, it's a bit weird}\antoinermk{It is the same color scheme (normals are in camera frame with COLMAP convention), it's just that the indoor scene has more diversity in the directions. I can try to make it look better}
% To evaluate the quality and completeness of exploration trajectories obtained with FisherRF~\cite{jiang2024fisherrf}, MACARONS~\cite{guedon2023macarons} and our method, we optimize for each method a 3D Gaussian splatting representation on 100 images collected along the trajectory.
% %
% Specifically, we apply Mesh-In-the-Loop Gaussian splatting~\cite{guedon2025milo} to allow for both novel view synthesis and surface mesh extraction. In each scene, the same starting viewpoint was used for all methods. 
% %
% The top row showcases renderings obtained with 3D Gaussians, and the bottom row presents normal maps of the extracted meshes.
%
% }
}
\vspace*{0mm}
\label{fig:qualitative_comparison}
\end{figure*} 
To the best of our knowledge, we are the first work that evaluates in both large-scale indoor and outdoor environments with varying action spaces.
Previous approaches are evaluated in either indoor or outdoor environments, and adapting many of these methods to the alternate setting is non-trivial.
\begin{table}
\centering
\setlength{\tabcolsep}{11pt}
\small
\caption{\textbf{Evaluation results on the Macarons++ dataset.} 
% All methods are evaluated by 100-step explorations from the same five randomly sampled initial camera poses per scene, using RGB-D inputs.
}
\label{tab:coverage_auc_real_world}
\begin{tabular}{lcc} \toprule
 & AUC $\uparrow$ & Final coverage $\uparrow$ \\ \midrule
Random Walk & 0.241 & 0.324 \\
SCONE~\cite{guedon2022scone} & 0.534 & 0.670 \\
MACARONS~\cite{guedon2023macarons} & 0.647 & 0.819 \\
FisherRF~\cite{jiang2024fisherrf} & 0.546 & 0.786 \\
MAGICIAN (Ours) & \textbf{0.721} & \textbf{0.919} \\ \midrule
\end{tabular}
\end{table}

\begin{table}[t]
\small
\tabcolsep=0.12cm
\caption{\textbf{Evaluation of novel-view rendering and mesh reconstruction on large-scale real-world scanned scenes.} Our method achieves the best performance across all metrics.}
\label{tab:mesh_novel_view}
\begin{tabular}{ccccc} \toprule
Method & SSIM $\uparrow$ & PSNR $\uparrow$ & LPIPS $\downarrow$ & Acc. (\%) $\uparrow$ \\ \midrule
FisherRF \cite{jiang2024fisherrf} & 0.55 & 13.95 & 0.38 & 79.15 \\
MACARONS \cite{guedon2023macarons} & 0.61 & 15.68 & 0.34 & 86.42 \\
\ourmodel{} (Ours) & \textbf{0.64} & \textbf{17.12} & \textbf{0.30} & \textbf{94.20} \\ \bottomrule
\end{tabular}
\end{table}

% \begin{table}[t]
% \centering
% \small  
% \setlength{\tabcolsep}{7.pt}  
% \scalebox{0.95}{
% \footnotesize
% \begin{tabular}{c|cccc}
% % \toprule
% Method & SSIM $\uparrow$ & PSNR $\uparrow$ & LPIPS $\downarrow$ & Acc. (\%) $\uparrow$ \\
% \hline
% FisherRF \cite{jiang2024fisherrf} & \tbest 0.55 & \tbest 13.95 & \tbest 0.38 & \tbest 79.15 \\
% MACARONS \cite{guedon2023macarons} & \sbest 0.61 & \sbest 15.68 & \sbest 0.34 & \sbest 86.42 \\
% Ours & \best {0.64} & \best {17.12} & \best {0.30} & \best {94.20} \\
% % \hline
% \end{tabular}}
% \caption{\textbf{Evaluation of novel-view rendering and mesh reconstruction on large-scale real-world scanned scenes.} Our method achieves the best performance across all metrics.}
% \label{tab:mesh_novel_view}
% \end{table}
\noindent\textbf{Macarons++ dataset.}
We benchmark against state-of-the-art methods: SCONE \cite{guedon2022scone}, MACARONS \cite{guedon2023macarons}, and FisherRF \cite{jiang2024fisherrf}\footnote{We adapt the released FisherRF code to outdoor scenes by modifying its frontier selection and adjusting its action space.}. 
\Cref{tab:coverage_auc_real_world} demonstrates that our method significantly outperforms all existing approaches in both reconstruction efficiency~(AUC) and final coverage, exceeding previous methods by a large margin.

SCONE and MACARONS, which adopt a greedy next-best-view strategy, perform well in simple outdoor environments. When applied to more complex or indoor scenes, their performance degrades significantly due to the lack of long-term planning, often causing the agent to be trapped in local regions before proceeding to explore new areas.

In contrast, FisherRF selects viewpoints along the frontier and generates a set of shortest paths from the current pose. It then evaluates these paths using Fisher Information to select the one with the highest expected information gain.
While this approach is effective for indoor active mapping, it relies heavily on frontier-based exploration and lacks global path optimization, leading to inefficient trajectory execution and unnecessary movement overhead. 
%
% The detailed table and more qualitative comparisons are included in the supplementary material.

To further evaluate the active mapping performance of our method, we conduct an additional comparison with MACARONS and FisherRF. For each scene and trajectory, we apply Mesh-in-the-Loop~(MILo) Gaussian Splatting~\cite{guedon2025milo} to 100 RGB-D frames collected during exploration, enabling both novel view synthesis and surface mesh reconstruction. Results in \Cref{tab:mesh_novel_view} demonstrate that the trajectories generated by \ourmodel{} also lead to better mesh reconstruction and novel-view synthesis. \Cref{fig:qualitative_results} shows our results, and \Cref{fig:qualitative_comparison} shows qualitative comparisons.

\begin{table}[t]
  \centering
  \caption{\textbf{Evaluation results on the MP3D dataset.} Our method consistently outperforms existing approaches under various robot and action-space settings.}
  \label{tab:mp3d_table}
  \setlength{\tabcolsep}{5pt}
  \scalebox{0.95}{
  \small
  \begin{tabular}{lccc}
    \toprule
    {Setting} & {Method} & {Comp. (\%) $\uparrow$} & {Comp. (cm) $\downarrow$} \\
    \midrule
    \multirow{7}{*}{\begin{tabular}[c]{@{}l@{}}Wheeled \\ Robot\end{tabular}}
      & Random & 45.67 & 26.53 \\
      & FBE \cite{yamauchi1997frontier} & 71.18 & 9.78 \\
      & UPEN \cite{georgakis2022uncertainty} & 69.06 & 10.60 \\
      & OccAnt \cite{ramakrishnan2020occupancy} & 71.72 & 9.40 \\
      & ANM \cite{yan2023active} &  73.15 &  9.11 \\
      & NBP \cite{li2025nextbestpath} &  79.38 &  6.78 \\
      & MAGICIAN (Ours) &  \textbf{85.45} &  \textbf{4.93} \\
    \midrule
    \multirow{3}{*}{Drone} 
      & NARUTO \cite{feng2024naruto} &  90.18 &  3.00 \\
      & ActiveGamer \cite{chen2025activegamer} &  95.32 &  2.30 \\
      & MAGICIAN( Ours) & \textbf{96.83} &  \textbf{2.11} \\
    \bottomrule
  \end{tabular}
  }
  
\end{table}

\noindent\textbf{MP3D dataset.}
We also compare our approach with state-of-the-art methods on the MP3D dataset.
As shown in \Cref{tab:mp3d_table}, even without further fine-tuning on MP3D dataset, our method outperforms existing approaches across different robot embodiments and action spaces. 
Moreover, we are the first to achieve state-of-the-art performance without relying on any traditional planner or a dedicated navigation model, thanks to our effective world modeling and beam search strategy.

\subsection{Ablation Study}

% We further demonstrate the strong performance and generalization capability of our method below. 
We conduct ablation studies on three unseen and challenging scenes: Sestino Museum, St. Sofia Church, and Neuschwanstein Castle of the Macarons++ dataset.

\begin{figure}[t]
  \centering
  \begin{subfigure}{0.5\columnwidth}
    \centering
    \includegraphics[width=\linewidth]{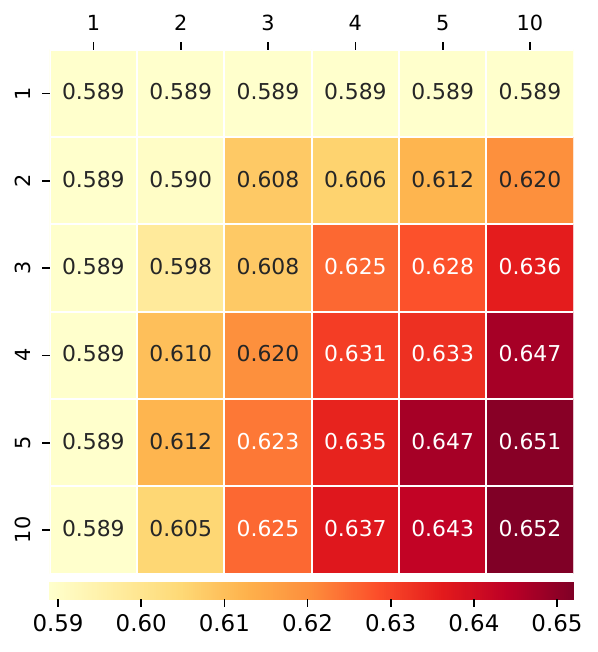}
    \caption{AUC}
    \label{fig:sub-a}
  \end{subfigure}\hfill
  \begin{subfigure}{0.5\columnwidth}
    \centering
    \includegraphics[width=\linewidth]{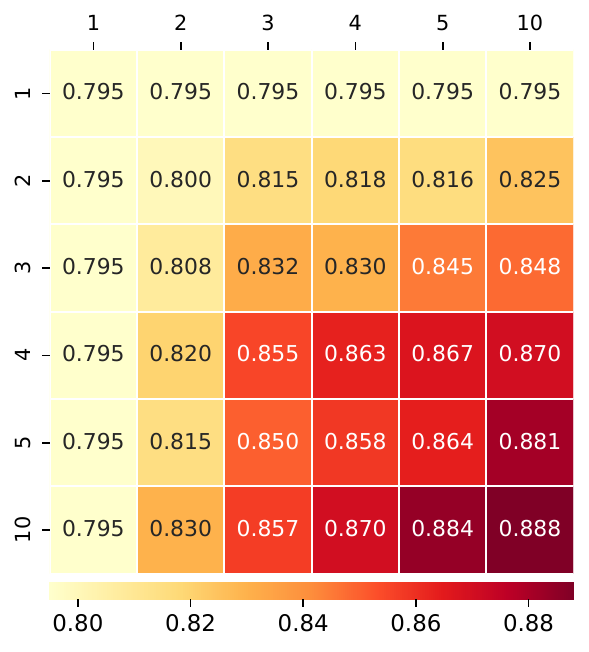}
    \caption{Final Coverage}
    \label{fig:sub-b}
  \end{subfigure}
  \caption{\textbf{Ablation study on the beam search parameters.} The horizontal axis denotes the beam width~$N_b$, and the vertical axis represents the look-ahead steps~$N_d$. Five steps correspond roughly to half the size of the scene.
  % \vincentrmk{I think we should still explain what 1 step correspond to in terms of distances, it could sound small to a reviewer}
  % \vincent{Five steps correspond roughly to the size of the scenes.} 
  % \shiyaormk{maybe we don't need to add this, The size and grid design of each scene are not uniformly consistent.}
  }
  % \vspace{-0.5cm}
  \label{fig:heatmap}
\end{figure}

\noindent
\textbf{Beam search.} \Cref{fig:heatmap} shows that the results consistently improve as the number of beams $N_b$ and the number of look-ahead steps $N_d$ increase, demonstrating the effectiveness of the proposed beam search strategy.
Increasing the number of beams or look-ahead steps yields an absolute improvement of 6.3\% in AUC and 9.3\% in final coverage. 

\noindent\textbf{Imagined Gaussians for coverage gain computation.}
When either the beam width or the look-ahead depth is set to 1, the method degenerates into a greedy next-best-view selection. Even in this case, our approach still surpasses MACARONS by 5.2\% in AUC and 10.9\% in final coverage, highlighting the advantage of using volumetric rendering with Imagined Gaussians for computing the coverage gain rather than using the Monte Carlo approximation of MACARONS.
Furthermore, we conducted a direct comparison of surface coverage gain computation efficiency with MACARONS. When evaluating a single candidate viewpoint under identical settings, our method achieves a $25\times$ speedup, requiring only $0.002$s compared to $0.05$s for MACARONS.

% \shizhermk{Fig~\ref{fig:imagined_gaussians} is not referred anywhere. You can add a small paragraph here. }

\begin{figure}[t]
  \centering
  \includegraphics[width=1\linewidth]{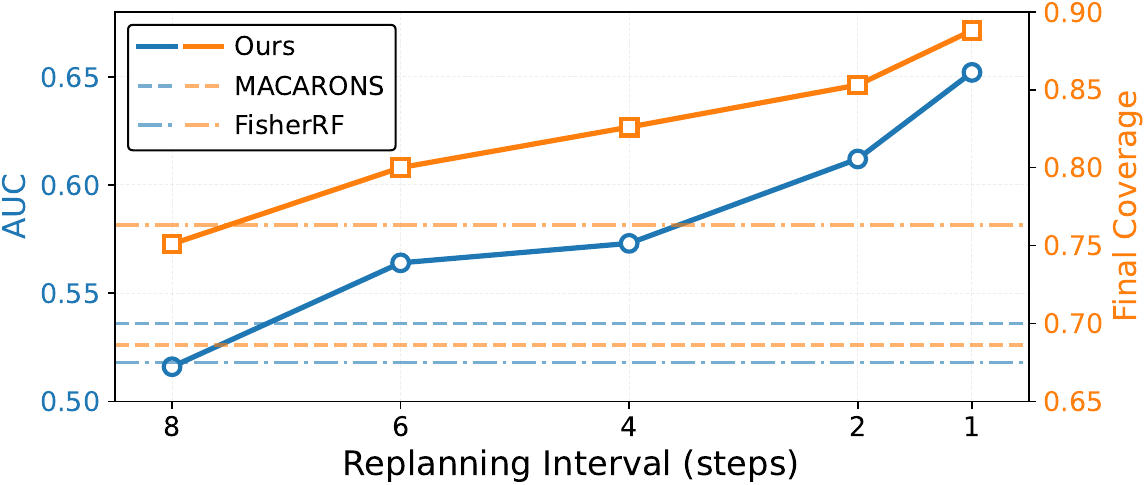}
  \caption{\textbf{Ablation study on replanning frequency.} The horizontal axis indicates the number $N_f$ of movement steps executed after each planning phase before replanning. 
  % Our performance improves with more frequent updates; however, replanning every 6 steps already provides state-of-the-art results.
  Replanning every 6 steps already provides state-of-the-art results.
  }
  \label{fig:replanning}
\end{figure}

\noindent
\textbf{Replanning frequency.} \Cref{fig:replanning} shows the performance improves with more frequent updates of the trajectory and occupancy predictions. 
However, we still obtain very good performance with less frequent replanning: Replanning every 6 steps already provides state-of-the-art results.
% }

% \textbf{Trajectory updates.} To evaluate the impact of 3D modeling accuracy on overall performance, we conduct experiments with different replanning intervals. At each replanning step, the imagined Gaussians is updated, and the number of executed movements equals the length of the replanning interval. As shown in \cref{fig:replanning}, our performance consistently improves with more frequent updates of the proxy model.

\noindent
\textbf{Fine-tuning occupancy model in indoor scenes.} To further verify that a strong occupancy model is not necessary to achieve good performance, we fine-tuned the occupancy model on the MP3D dataset and evaluated on these three scenes. 
%\cref{tab:ablation_finetuning} shows a trade-off between exploration efficiency and final coverage, with only a marginal improvement (0.5\%) in final coverage after fine-tuning.
\cref{tab:ablation_finetuning} shows that fine-tuning on indoor scenes does not provide any clear improvement.

We present additional tables and more qualitative comparisons in the supplementary material.
%
% \begin{table}[t]
%   \centering
%   \setlength{\tabcolsep}{23pt}
%   \scalebox{0.99}{
%   \footnotesize
%   \begin{tabular}{c|cc}
%     % \toprule
%     Occupancy Model & AUC & Cov. \\
%     \hline
%     Pretrained  & \best {0.652} & \sbest 0.888 \\
%     Fine-tuned & \sbest 0.646 & \best {0.893} \\
%     % \hline
%   \end{tabular}
%   }
%   \caption{
%   \textbf{Ablation study on comparing models with and without fine-tuning on indoor environments.} The fine-tuned version shows a minor 0.5\% improvement in final coverage, while the original model retains higher exploration efficiency.
%   % \todo{Move to suppmat.}
%   }
%   \label{tab:ablation_finetuning}
% \end{table}
\begin{table}[t]
  \centering
  \setlength{\tabcolsep}{28pt}
  \caption{
  \textbf{Ablation study on comparing models with and without fine-tuning on indoor environments.} The fine-tuned version shows a minor 0.5\% improvement in final coverage, while the original model retains higher exploration efficiency.
  % \todo{Move to suppmat.}
  }
  \label{tab:ablation_finetuning}
  \small
  \begin{tabular}{@{}ccc@{}} \toprule
    % \toprule
    Occupancy Model & AUC $\uparrow$ & Cov. $\uparrow$ \\ \midrule
    Pretrained  &  \textbf{0.652} &  0.888 \\
    Fine-tuned &  0.646 &  \textbf{0.893} \\ \bottomrule
    % \hline
  \end{tabular}
  
\end{table}

\section{Conclusion}
% \shizhermk{This needs to be improved.}
% In this paper, we demonstrated the importance of long-term planning for the efficient exploration of complex scenes. Our approach performs long-term planning by applying beam search over future possible moves and optimizing for predicted surface coverage. The efficiency of the beam search is enabled by our volumetric representation, which allows rapid estimation of coverage gain. By combining learning-based components (the prediction of a probabilistic occupancy volume) with non-learning-based components (the beam search), our method achieves strong performance across diverse types of scenes.
In this paper, we addressed the long-standing challenge of efficient active mapping by introducing \ourmodel{}, a framework that models the world from past observations to plan future exploration. By combining a pre-trained probabilistic occupancy network with a volumetric Imagined Gaussian representation, our method enables fast estimation of coverage gain and efficient beam-search–based long-term planning, achieving superior performance across diverse indoor and outdoor scenes. 
% \shizhermk{shorten the future work if spaces are limited.}
Looking ahead, the rise of 3D foundation models~\cite{wang2024dust3r, wang2025vggt} opens new opportunities to extend active mapping toward purely RGB-based exploration without relying on depth or pose information. Furthermore, incorporating semantic~\cite{chen2025understanding}  could enable more informative and goal-directed exploration.

\section*{Acknowledgements}

This project was funded by the European Union (ERC Advanced Grant explorer Funding ID \#101097259) and the ANR project 3D-GEM ANR-25-CE23-7777-01. This work was granted access to the HPC resources of IDRIS under the allocation 2025-AD011014703R2 made by GENCI. We thank Hongyu Zhou for his valuable help during the experimental evaluation phase of this work.
{
    \small
    \bibliographystyle{ieeenat_fullname}
    \bibliography{main}
}

% WARNING: do not forget to delete the supplementary pages from your submission 
\clearpage
\setcounter{page}{1}
% \maketitlesupplementary
\appendix
\section*{Appendix}
\begin{table*}[t]
\caption{\textbf{AUCs of full scenes on the Macarons++ dataset.}}
  \centering
  \setlength{\tabcolsep}{11pt}
  \small
  \begin{tabular}{l@{\hspace{1.5em}}ccccc}
    \toprule
    {Scene} & {Rand. Walk} & {SCONE~\cite{guedon2022scone}} & {MACARONS\cite{guedon2023macarons}} & {FisherRF~\cite{jiang2024fisherrf}} & \textbf{MAGICIAN (Ours)} \\
    % \midrule
    \hline
    Dunnottar Castle  & 0.149 & 0.366 & \sbest 0.618 & \tbest 0.500 & \best 0.745 \\
    Colosseum  & 0.219 & \tbest 0.589 & \sbest 0.656 & 0.551 & \best 0.704 \\
    Bannerman Castle  & 0.192 & 0.559 & \tbest 0.575 & \sbest 0.595 & \best 0.761 \\
    Pantheon  & 0.198 & \tbest 0.465 & \sbest 0.601 & 0.270 & \best 0.644 \\
    Christ the Redeemer  & 0.439 & \tbest 0.772 & \best 0.859 & 0.727 & \sbest 0.793 \\
    Statue of Liberty  & 0.323 & \tbest 0.632 & \sbest 0.711 & 0.553 & \best 0.797 \\
    Pisa Cathedral  & 0.290 & \tbest 0.486 & \sbest 0.678 & \tbest 0.486 & \best 0.723 \\
    Fushimi Castle  & 0.279 & \tbest 0.689 & \sbest 0.718 & 0.565 & \best 0.766 \\
    % \midrule
    \hline
    Alhambra Palace  & 0.126 & 0.369 & \sbest 0.567 & \tbest 0.462 & \best 0.631 \\
    Neuschwanstein Castle  & 0.184 & 0.325 & \sbest 0.452 & \tbest 0.375 & \best 0.608 \\
    Eiffel Tower  & 0.333 & \tbest 0.683 & \sbest 0.709 & 0.616 & \best 0.754 \\
    Manhattan Bridge  & 0.258 & 0.632 & \best 0.750 & \tbest 0.637 & \sbest 0.705 \\
    % \midrule
    \hline
    St. Sofia Church  & 0.280 & 0.532 & \sbest 0.621 & \tbest 0.608 & \best 0.710 \\
    Barts  & 0.214 & 0.551 & \tbest 0.660 & \sbest 0.673 & \best 0.831 \\
    Sestino Museum  & 0.132 & 0.367 & \tbest 0.537 & \sbest 0.571 & \best 0.637 \\
    % \midrule
    \hline
    {Average}  & {0.241} & {0.534} & \sbest {0.647} & \tbest {0.546} & \best {0.721} \\
    % \bottomrule
    \hline
  \end{tabular}
  
  \label{tab:supp_macarons_auc}
\end{table*}
\begin{table*}[t]
\caption{\textbf{Final Coverages of full scenes on the Macarons++ dataset.}}
  \centering
  \setlength{\tabcolsep}{11pt}
  \small
  \begin{tabular}{l@{\hspace{1.5em}}ccccc}
    \toprule
    \textbf{Scene} & {Rand. Walk} & {SCONE~\cite{guedon2022scone}} & {MACARONS~\cite{guedon2023macarons}} & {FisherRF~\cite{jiang2024fisherrf}} & \textbf{MAGICIAN (Ours)} \\
    %\midrule
    \hline
    Dunnottar Castle  & 0.225 & 0.527 & \sbest 0.820 & \tbest 0.809 & \best 0.975 \\
    Colosseum  & 0.272 & 0.755 & \sbest 0.794 & \tbest 0.757 & \best 0.872 \\
    Bannerman Castle  & 0.240 & 0.722 & \sbest 0.834 & \tbest 0.801 & \best 0.917 \\
    Pantheon  & 0.309 & \tbest 0.610 & \sbest 0.796 & 0.444 & \best 0.842 \\
    Christ the Redeemer  & 0.581 & \tbest 0.924 & \sbest 0.967 & 0.876 & \best 0.973 \\
    Statue of Liberty  & 0.443 & \tbest 0.819 & \sbest 0.909 & \tbest 0.819 & \best 0.947 \\
    Pisa Cathedral  & 0.353 & 0.566 & \sbest 0.865 & \tbest 0.776 & \best 0.941 \\
    Fushimi Castle  & 0.449 & \tbest 0.844 & \sbest 0.853 & 0.814 & \best 0.931 \\
    % \midrule
    \hline
    Alhambra Palace  & 0.162 & 0.473 & \sbest 0.775 & \tbest 0.615 & \best 0.852 \\
    Neuschwanstein Castle  & 0.223 & 0.444 & \tbest 0.551 & \sbest 0.582 & \best 0.848 \\
    Eiffel Tower  & 0.541 & \tbest 0.856 & \sbest 0.915 & 0.827 & \best 0.923 \\
    Manhattan Bridge  & 0.356 & 0.781 & \sbest 0.924 & \tbest 0.877 & \best 0.955 \\
    % \midrule
    \hline
    St. Sofia Church  & 0.331 & 0.619 & \tbest 0.795 & \sbest 0.865 & \best 0.891 \\
    Barts  & 0.240 & 0.677 & \tbest 0.768 & \sbest 0.878 & \best 0.996 \\
    Sestino Museum  & 0.141 & 0.430 & \tbest 0.713 & \sbest 0.842 & \best 0.924 \\
    % \midrule
    \hline
    {Average}  & {0.324} & {0.670} & \sbest {0.819} & \tbest {0.786} & \best {0.919} \\
    %\bottomrule
    \hline
  \end{tabular}
  
  \label{tab:supp_macarons_cov}
\end{table*}

In \cref{sec:method_supp}, we present the details of the occupancy module, and the complete formulation of the coverage gain computation along with the analytical derivation of the depth-dependent weighting. In \cref{sec:supp_exp}, we provide additional implementation details, detailed tables, additional quantitative comparisons, and additional ablation studies. In \cref{sec:failure}, we discuss observed failure cases and provide an analysis.

\section{Method}
\label{sec:method_supp}

\subsection{Neural Occupancy Prediction}

Here, we provide additional architectural details of the volume occupancy 
module $\occ(\vecx \mid \bC_t)$. 

At each time step $t$, the occupancy 
module receives a 3D query point $\vecx$, the 
reconstructed surface point cloud $S_t$, and the previously visited 
camera poses $\bC_t$, and predicts an occupancy value in $[0,1]$ for $\vecx$.

To capture the local geometry around $\vecx$, we compute its 
$k$-nearest neighbors in $S_t$ and encode this neighborhood using a 
self-attention unit followed by pooling. To capture larger-scale 
structure, we repeat this procedure on progressively downsampled versions of $S_t$: at each scale, we recompute the neighbors of $\vecx$ 
and process them with an additional self-attention–pooling block. Coarser 
scales naturally expand the receptive field, allowing the model to 
integrate fine-grained and global geometric information.

The multi-scale features are concatenated and fed into an MLP to predict the occupancy value $\occ(\vecx \mid \bC_t)$. Because the architecture operates solely on local neighborhoods at each scale, it can be applied efficiently to large point clouds while still preserving fine geometric details. In practice, we set $k = 16$ and use three neighborhood scales.

We adopt this model architecture from \cite{guedon2023macarons} without modification. The diagram of this model architecture is presented in Figure 7 of that work.

\subsection{Coverage Gain Computation}
\label{sec:supp_coverage_gain}

\subsubsection{Coverage Gain Formulation}

For each candidate camera pose $\cam$, we compute the coverage gain $G_{\text{rendered}}(\cam)$ by rendering depth and novelty maps from the current Imagined Gaussian state using volumetric rendering (Eq.~\eqref{eq:volumetric-rendering-equation-with-coverage} in the main paper):
\begin{equation}
G_{\text{rendered}}(\cam)
= \sum_{\pix \in \mathcal{P}_{\text{valid}}}
w_{\text{depth}}(\pix)\, I_{\text{novelty}}(\pix),
\label{coverage-gain-weight}
\end{equation}
where $\mathcal{P}_{\text{valid}} = \{\pix \mid D(\pix) > 0\}$ denotes pixels with valid depth $D(\pix)$, $I_{\text{novelty}}(\pix)$ is the rendered novelty value, and $w_{\text{depth}}(\pix)$ is a depth-dependent weighting factor:
\begin{equation}
w_{\text{depth}}(\pix)
= \min\left(1,\left(\frac{D(\pix)}{D_{\text{th}}}\right)^2\right),
\label{eq:depth_weight}
\end{equation}
where $D_{\text{th}}$ denotes a threshold and is set to half of the estimated scene scale. This weighting term mitigates oversampling at close range, where the pixel sampling density of the depth sensor exceeds the resolution required for faithful surface reconstruction.

\subsubsection{Analytical Derivation of Depth Weighting}

\textbf{Target surface density.} 
Surface coverage becomes well defined only after specifying a target spatial resolution. For large urban scenes, one point per square decimeter may suffice, whereas tabletop objects typically require several points per square centimeter. We denote this desired sampling resolution as the \emph{target surface density} $r_{\text{target}}$, representing the minimum number of points per unit surface area required for adequate reconstruction. This concept is commonly used in existing methods~\cite{guedon2022scone, guedon2023macarons}.

Since the depth sensor always captures a fixed number of samples $N = H \times W$ per frame, the local surface sampling density depends solely on the distance between the camera and the observed surface. By Thales's theorem, this density decays quadratically with depth. Consequently, when the sensor is too close to a surface, the resulting sample density exceeds $r_{\text{target}}$ and provides no additional benefit for coverage. Thus, $r_{\text{target}}$ naturally induces a threshold depth $D_{\text{th}}$, below which moving the camera closer becomes inefficient.

\textbf{Mathematical derivation.}
Consider a square patch of the depth map with side length $s$ centered at pixel $\pix$. This patch contains
\begin{equation}
n_{\text{captured}} = s^2
\end{equation}
captured depth samples. By Thales's theorem, the corresponding 3D surface region has area:
\begin{equation}
A = \left(\frac{s\,D(\pix)}{f}\right)^2,
\end{equation}
where $f$ is the focal length in pixel units. The resulting surface sampling density is therefore:
\begin{equation}
r(\pix)
= \frac{n_{\text{captured}}}{A}
= \left(\frac{f}{D(\pix)}\right)^2,
\end{equation}
confirming the inverse-square relationship with depth. The depth at which $r(\pix)$ equals the target density $r_{\text{target}}$ is obtained by solving $r(\pix) = r_{\text{target}}$:
\begin{equation}
D_{\text{th}} = \frac{f}{\sqrt{r_{\text{target}}}}.
\end{equation}

For depths $D(\pix) < D_{\text{th}}$, the captured sample density is unnecessarily high. In this regime, although the patch contains $n_{\text{captured}} = s^2$ samples, only $A \, r_{\text{target}}$ samples are needed to meet the target surface density. The fraction of samples that meaningfully contribute to coverage is thus:
\begin{equation}
p(\pix)
= \frac{A\, r_{\text{target}}}{n_{\text{captured}}}
= \frac{r_{\text{target}}}{r(\pix)}
= \left(\frac{D(\pix)}{D_{\text{th}}}\right)^2.
\end{equation}

Pixels observed at depths smaller than $D_{\text{th}}$ should therefore contribute only proportionally to $p(\pix)$, reflecting the redundancy introduced by oversampling in this regime.

Conversely, when $D(\pix) \ge D_{\text{th}}$, the sampling density satisfies $r(\pix) \leq r_{\text{target}}$, meaning that all captured samples are necessary and should contribute fully. Combining both regimes yields the depth-dependent weighting function:
\begin{equation}
w_{\text{depth}}(\pix)
= \min\left(1,\left(\frac{D(\pix)}{D_{\text{th}}}\right)^2\right).
\end{equation}

This weighting strategy, used in Eq.~\eqref{coverage-gain-weight}, prevents the planner from favoring near-surface viewpoints that artificially inflate point counts without improving effective surface coverage. As a result, the exploration process is guided toward trajectories that yield more efficient and informative observations.

\section{Experiments}
\label{sec:supp_exp}

\subsection{Implementation Details}
Our simulation is built on PyTorch3D~\cite{pytorch3d}, which supports differentiable rendering and ray casting to generate RGB-D data from arbitrary camera viewpoints. The pretrained occupancy model was trained using four NVIDIA Tesla V100 SXM2 32 GB GPUs, while inference was performed on a single V100 GPU.

In our experiments on the Macarons++ dataset, we evaluated Final Coverage and AUC scores using ground-truth point clouds. However, unlike prior work~\cite{guedon2023macarons, li2025nextbestpath} that directly samples point clouds from the ground-truth mesh, which may include invisible points (e.g., points inside Pisa Cathedral), we generated the ground-truth point cloud by rendering depth maps from all accessible viewpoints and projecting them into a 3D point cloud.

For each scene, we evaluate 15 novel views. To obtain a set of novel views that cover the entire ground-truth mesh, we use a submodular optimization–based selection procedure. At each iteration, we randomly sample 100 candidate 6D poses within the scene’s bounding box and, for each pose, count how many ground-truth points are visible from that viewpoint. We then select the pose that observes the largest number of previously unseen ground-truth points and mask out those newly observed points from the ground-truth point cloud. We repeat this process by sampling a new batch of 100 candidate poses and again selecting the pose that reveals the most remaining unseen points, until 15 novel views are selected.

\subsection{Comparison with State-of-the-Art Methods}
In this section, we provide detailed evaluation results on the Macarons++ dataset, along with additional qualitative comparisons and analyses.

From \cref{tab:supp_macarons_auc} and \cref{tab:supp_macarons_cov}, we observe that the state-of-the-art NBV-based method MACARONS~\cite{guedon2023macarons} remains a very strong baseline in relatively simple scenes such as Manhattan Bridge and Christ the Redeemer. However, due to its lack of long-term planning, it struggles to escape already fully explored local regions, which leads to poor performance in indoor environments. FisherRF~\cite{jiang2024fisherrf}, which relies on frontier detection and Fisher information, performs reasonably better in indoor environments due to its frontier-based exploration. However, the frontier mechanism also introduces unnecessary movements, leading to inefficient trajectories, particularly in outdoor scenes. In contrast, our method is neither restricted by frontier heuristics nor hampered by short-sighted planning. By performing the tree search to identify full trajectories that maximize coverage gain, our method achieves state-of-the-art performance in both indoor and outdoor scenes.

\begin{figure}[t]
  \centering
  \includegraphics[width=0.92\linewidth]{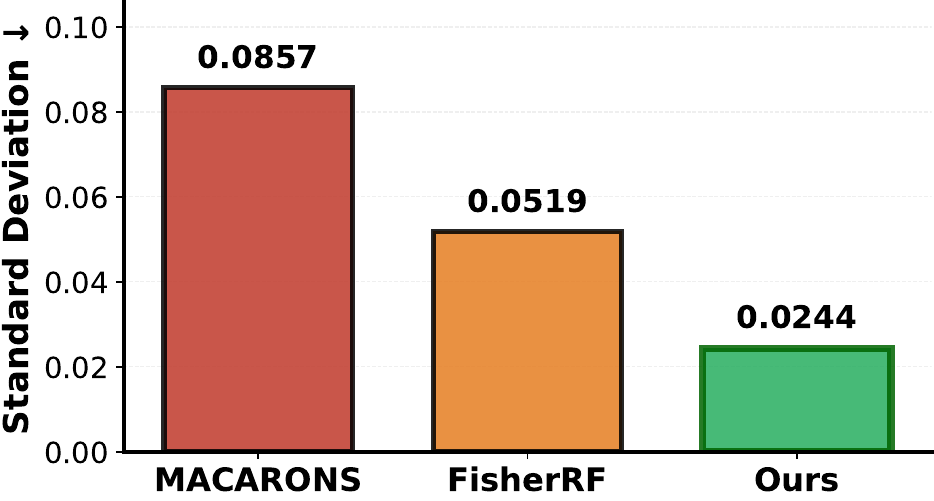}
  \caption{\textbf{Standard deviation of the final coverage across different methods and scenes.} Our method achieves consistently low values for this metric, indicating strong robustness to random starting poses, whereas other methods exhibit much larger variability.
  }
  \label{fig:se_std}
\end{figure}

As we mentioned in the main paper, during the evaluation stage, the five starting poses in each scene are randomly sampled. To more rigorously evaluate the stability of each method under this randomness, we compute the standard deviation of the final coverage for each method in each scene, and further compute their average across all scenes to summarize the overall variability. The results shown in \cref{fig:se_std} demonstrate that our method exhibits consistently low values in this metric, indicating that its performance is highly robust: despite different random initial poses, it reliably achieves high final coverage. In contrast, the other methods exhibit substantially larger variance, suggesting that their performance is highly sensitive to the initial pose and the corresponding early observations.

\begin{figure*}

\setlength{\fboxsep}{0pt}   % padding
\setlength{\fboxrule}{1.pt}% border thickness

\vspace*{7mm}

\centering

% FisherRF
\begin{minipage}[b]{0.33\textwidth}
    \centering
    {\includegraphics[width=\linewidth]{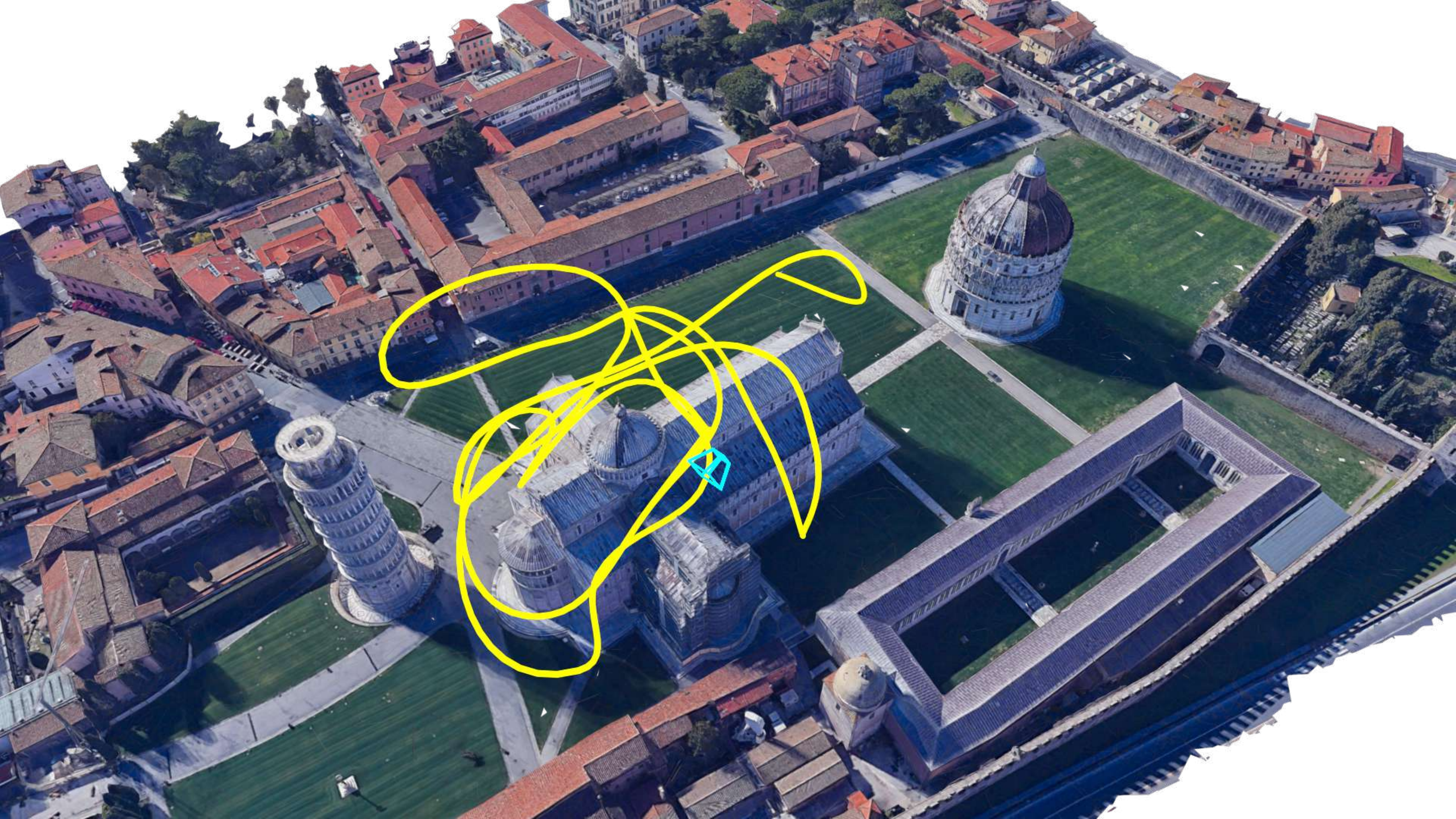}}
    {\includegraphics[width=\linewidth]{sec/images/render/pisa_fisherf_cam0/rgb.jpg}}
    {\includegraphics[width=\linewidth]{sec/images/render/pisa_fisherf_cam0/mesh_normal.jpg}}
    {\includegraphics[width=\linewidth]{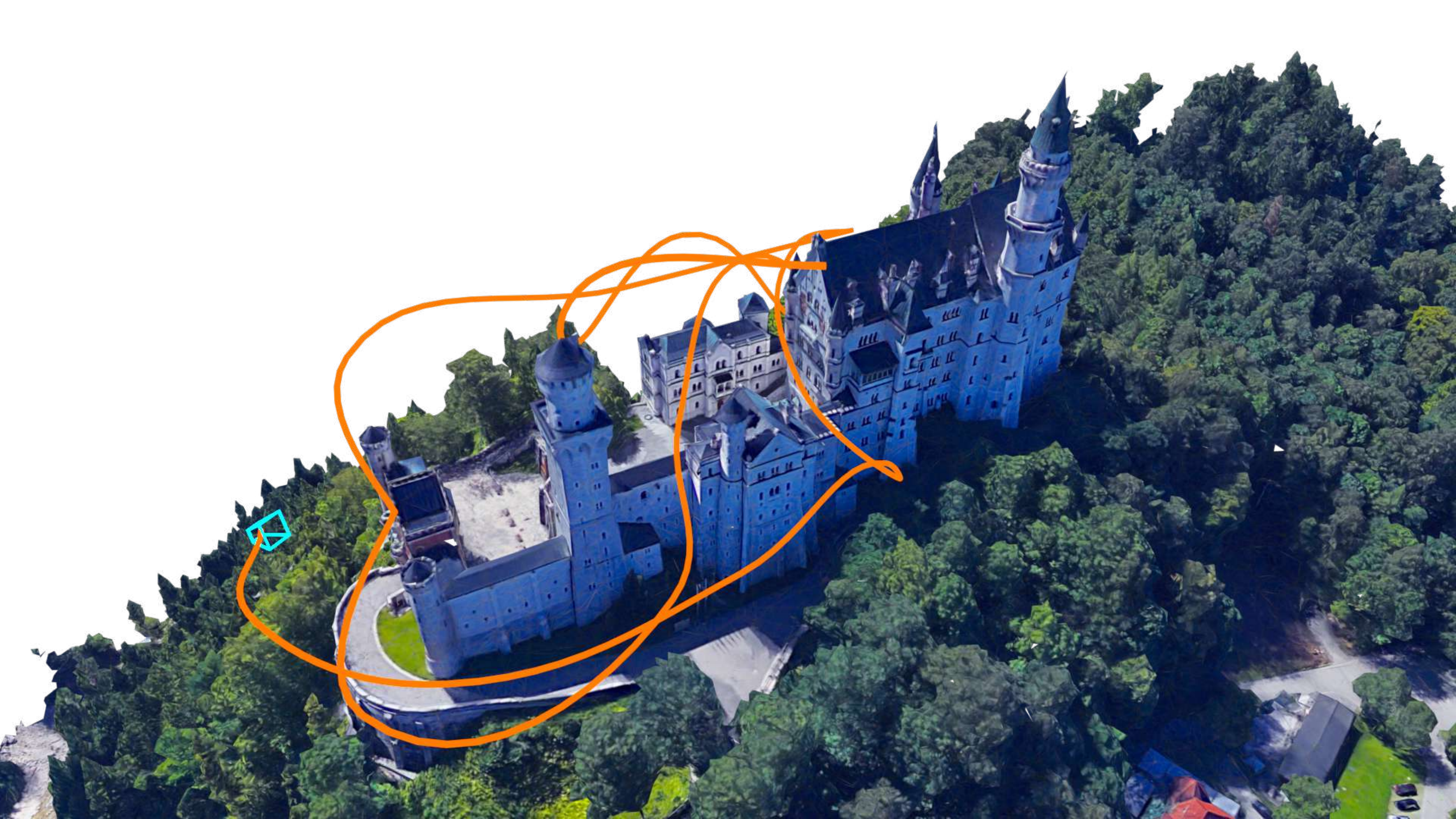}}
    {\includegraphics[width=\linewidth]{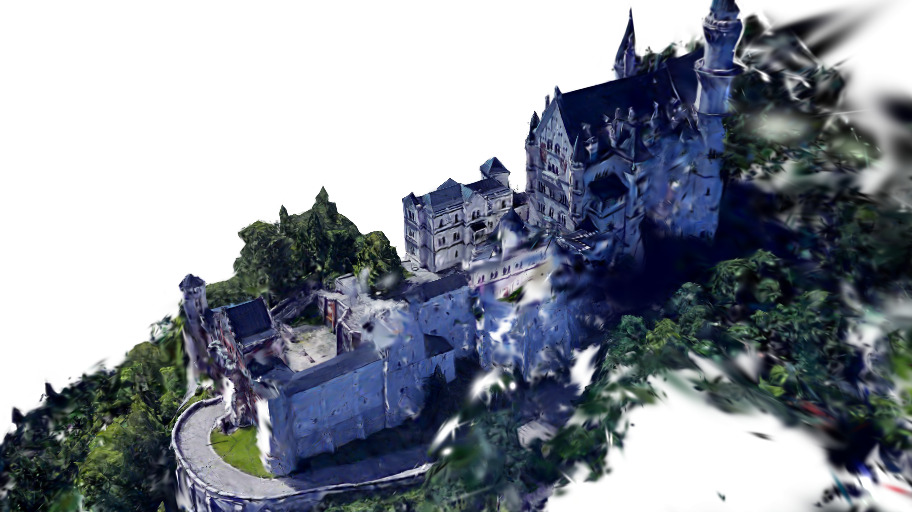}}
    {\includegraphics[width=\linewidth]{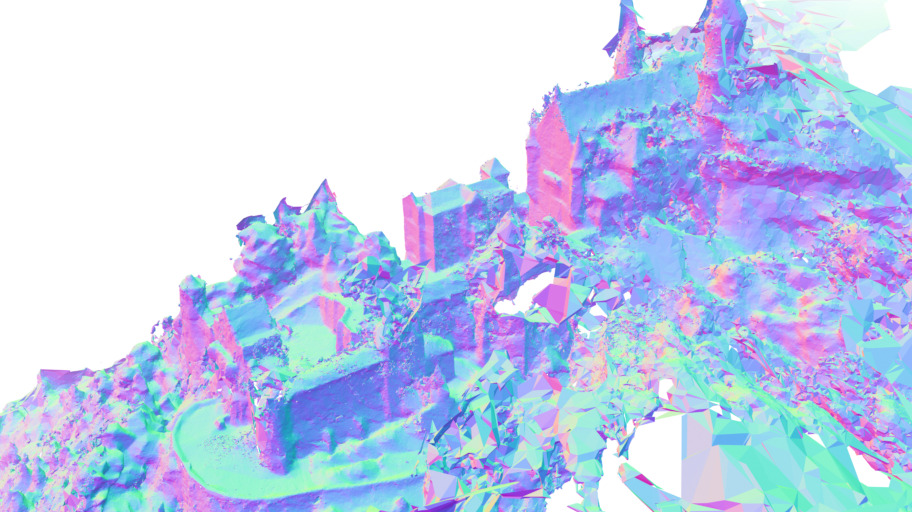}}
    \vfill
    FisherRF~\cite{jiang2024fisherrf}
\end{minipage}
\hfill
% MACARONS
\begin{minipage}[b]{0.33\textwidth}
    \centering
    {\includegraphics[width=\linewidth]{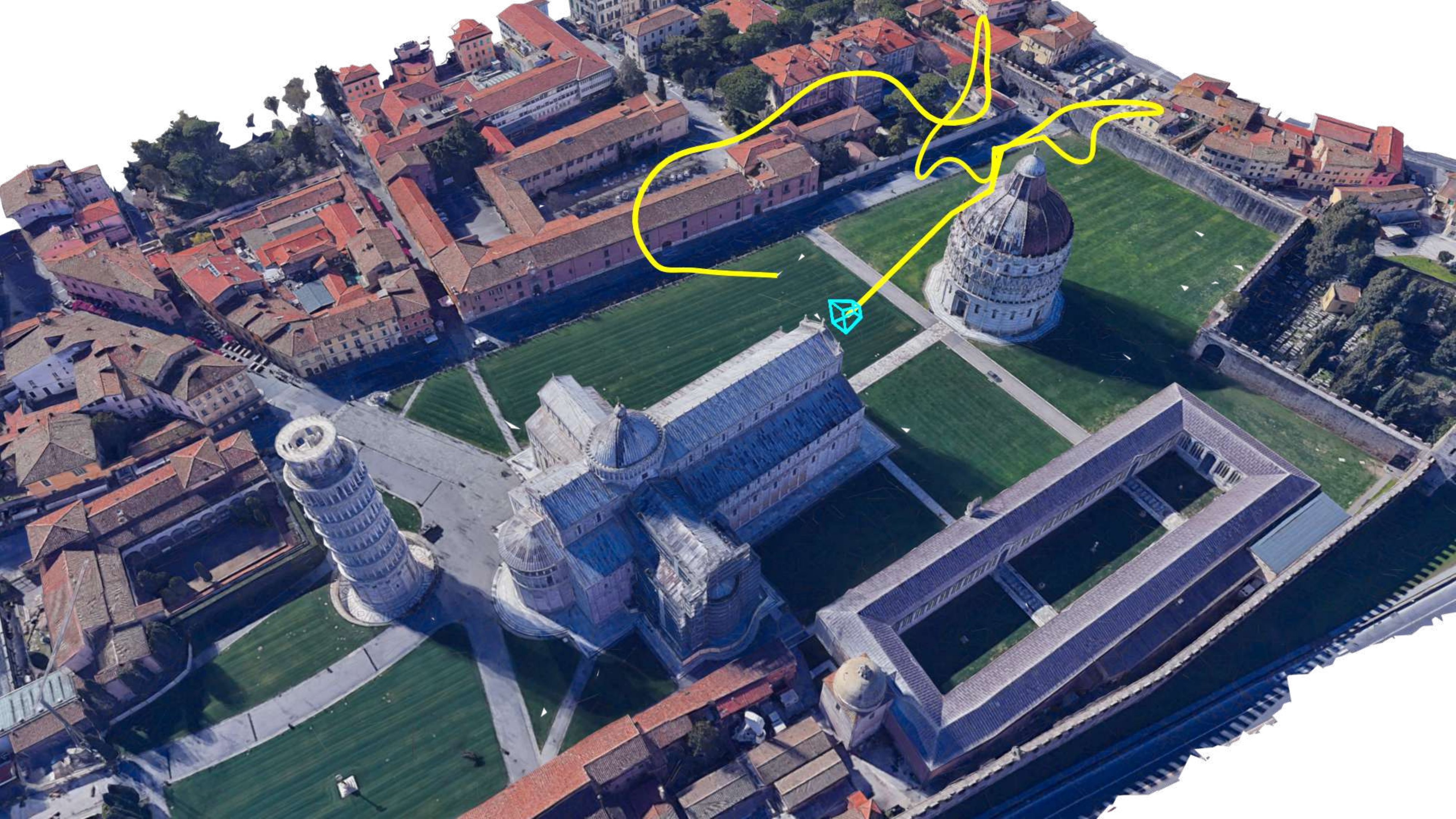}}
    {\includegraphics[width=\linewidth]{sec/images/render/pisa_macarons_cam0/rgb.jpg}}
    {\includegraphics[width=\linewidth]{sec/images/render/pisa_macarons_cam0/mesh_normal.jpg}}
    {\includegraphics[width=\linewidth]{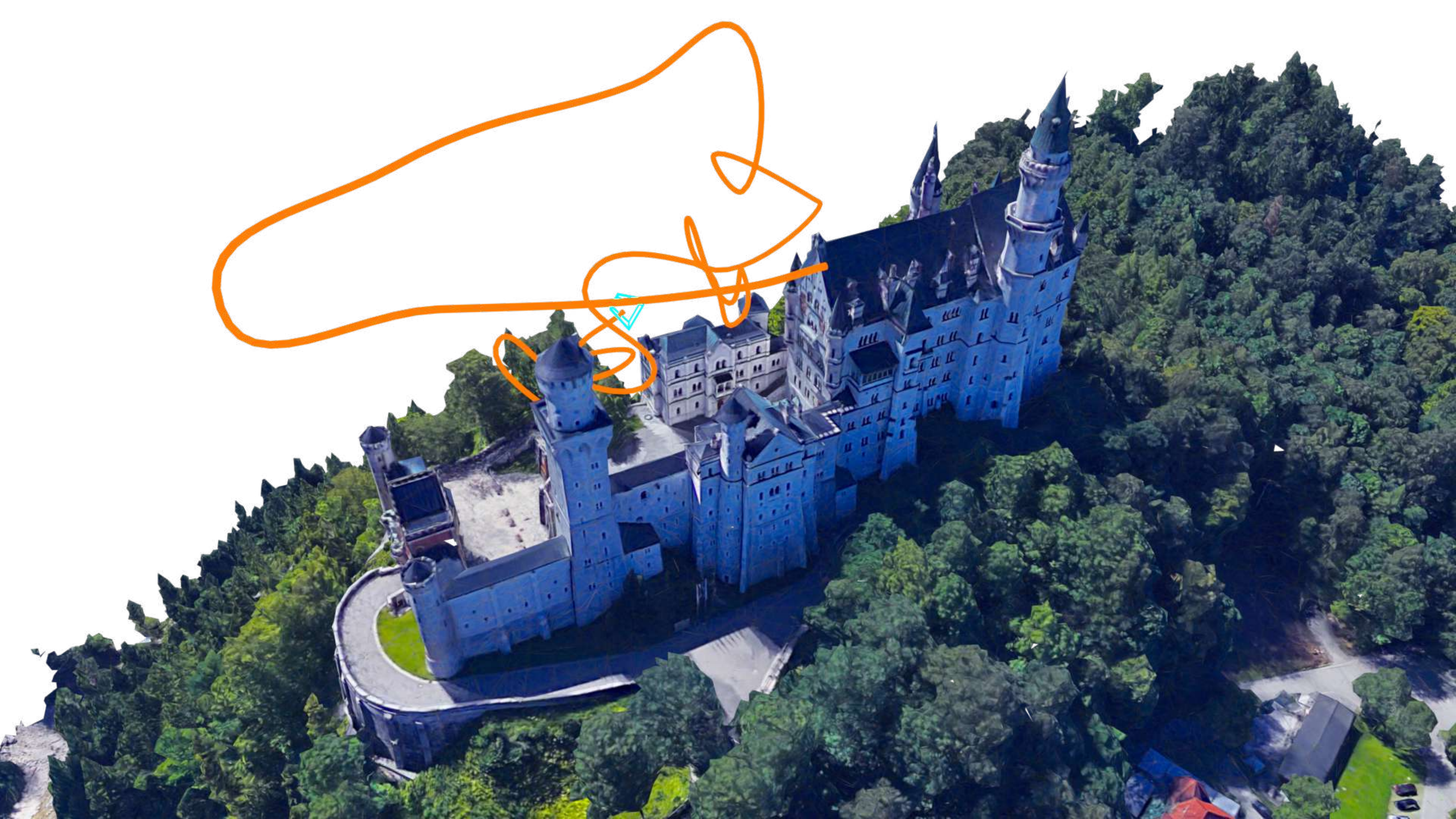}}
    {\includegraphics[width=\linewidth]{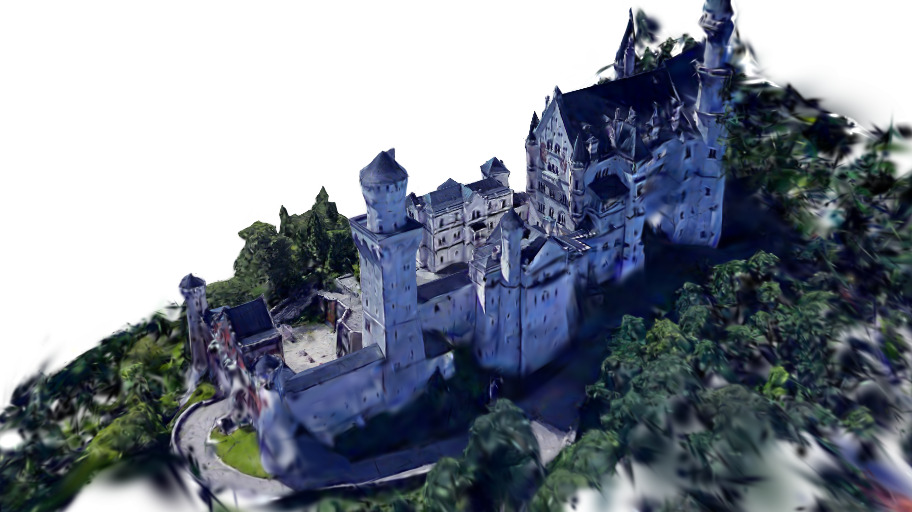}}
    {\includegraphics[width=\linewidth]{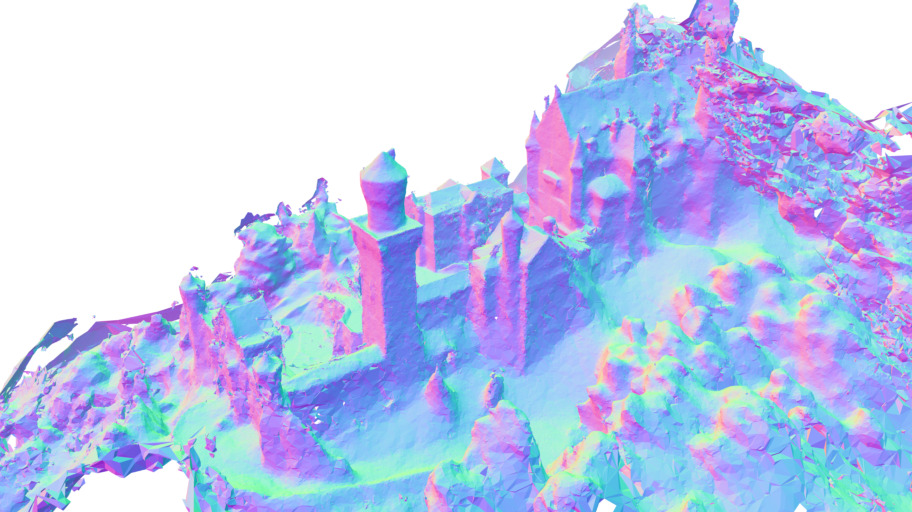}}
    \vfill
    MACARONS~\cite{guedon2023macarons}
\end{minipage}
\hfill
% Ours
\begin{minipage}[b]{0.33\textwidth}
    \centering
    {\includegraphics[width=\linewidth]{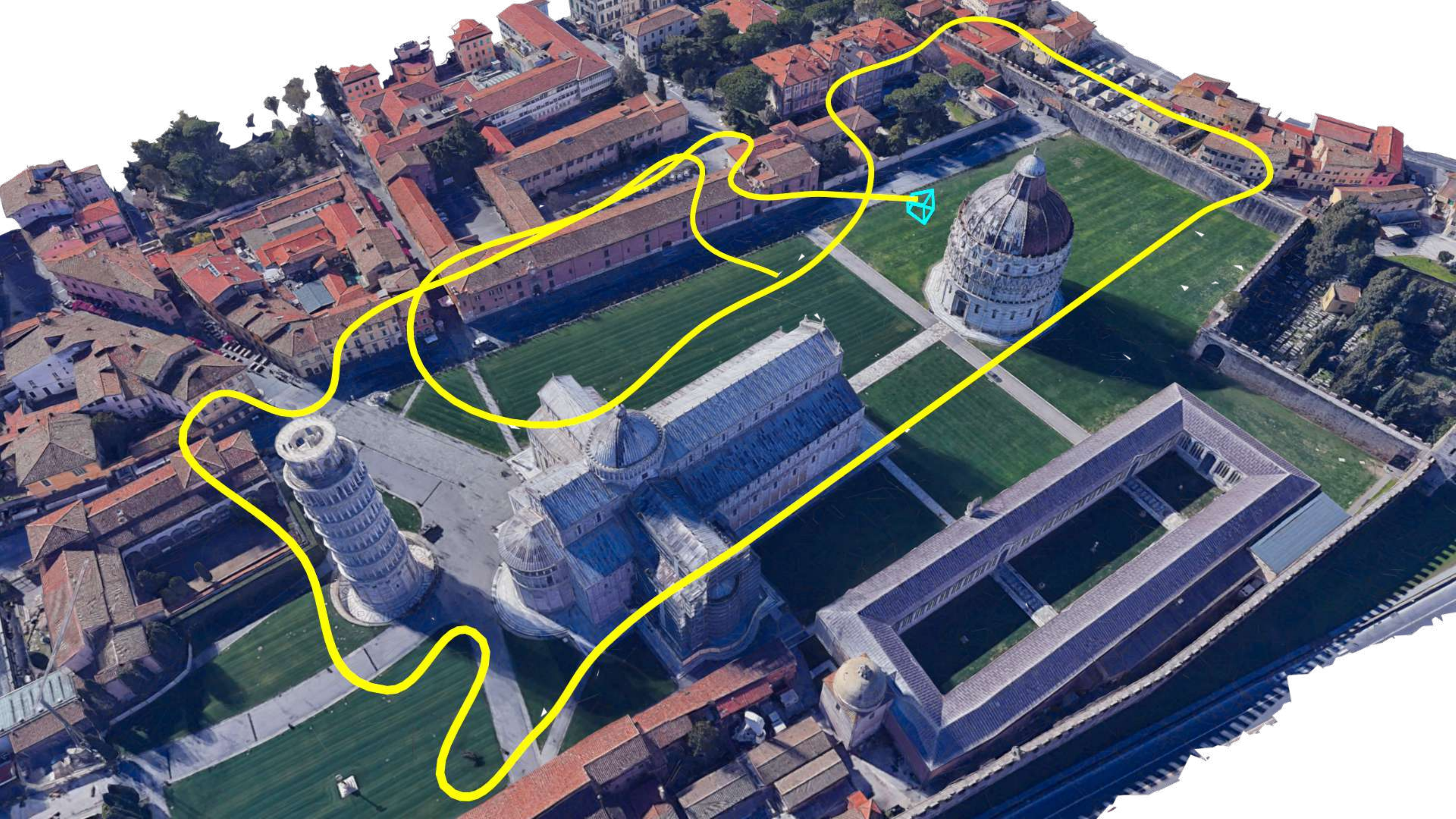}}
    {\includegraphics[width=\linewidth]{sec/images/render/pisa_ours1_cam0/rgb.jpg}}
    {\includegraphics[width=\linewidth]{sec/images/render/pisa_ours1_cam0/mesh_normal.jpg}}
    {\includegraphics[width=\linewidth]{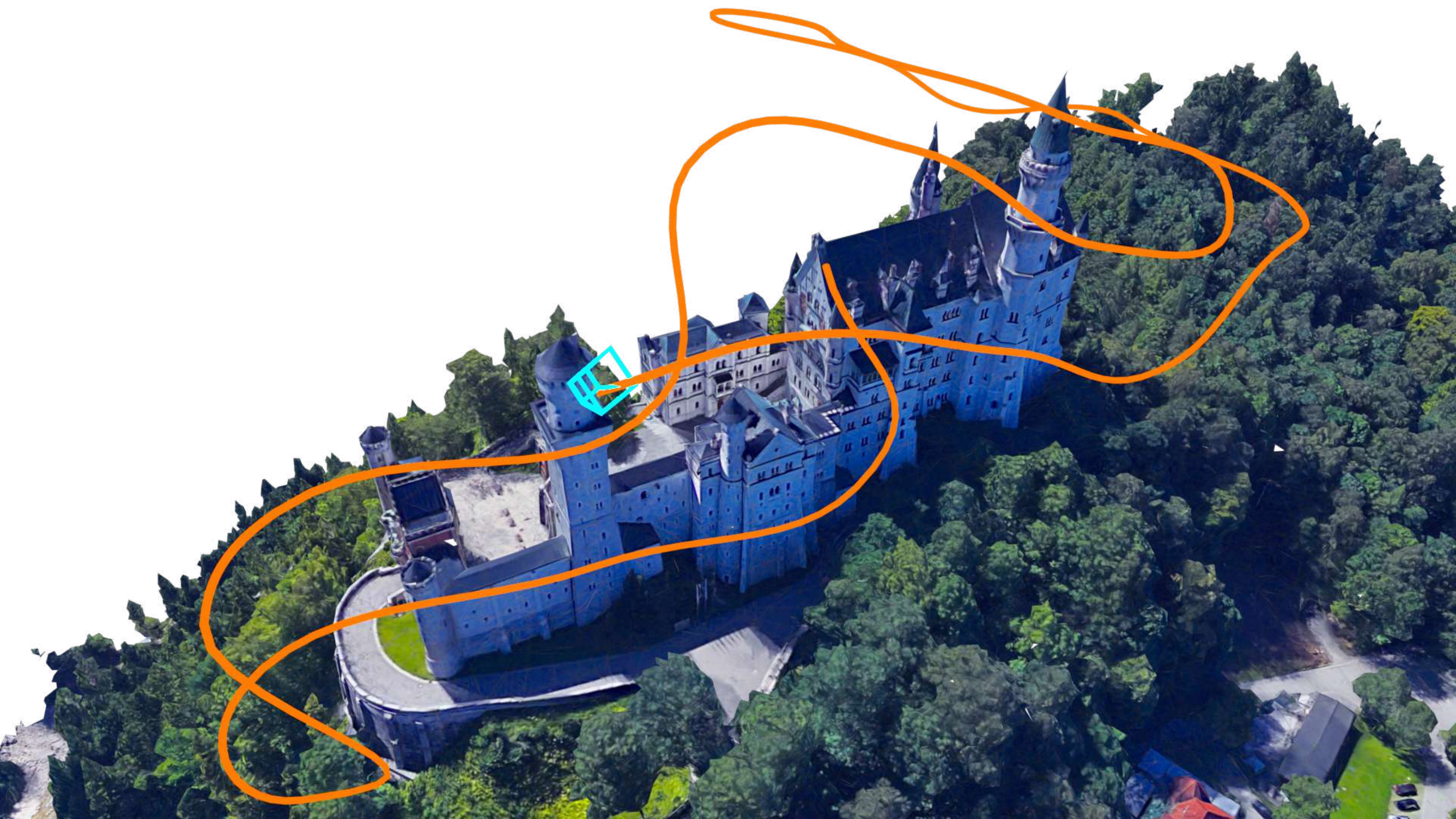}}
    {\includegraphics[width=\linewidth]{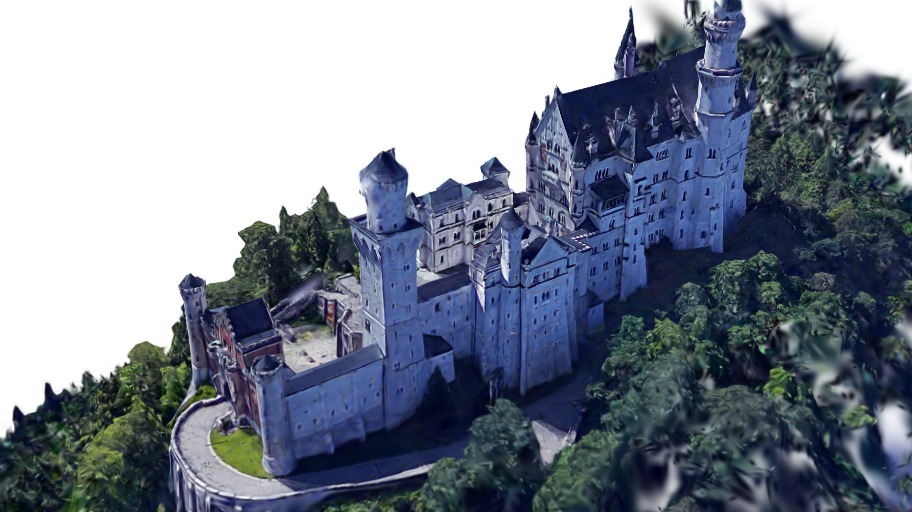}}
    {\includegraphics[width=\linewidth]{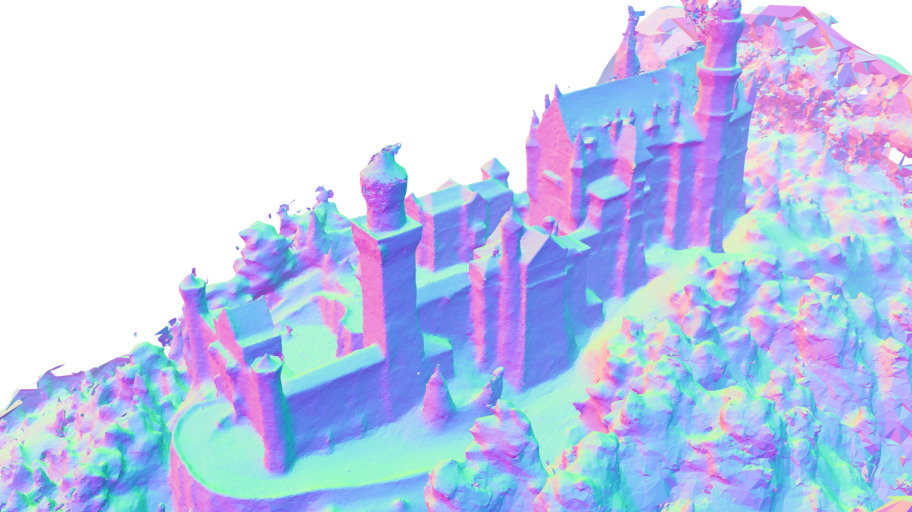}}
    \vfill
    Ours
\end{minipage}

\caption{
% \vincent{
\textbf{Visualization of exploration trajectories and qualitative comparisons of novel view synthesis and surface reconstruction in outdoor scenes. From top to bottom, the scenes are Pisa Cathedral and Neuschwanstein Castle.} 
In the same scene, all methods start from the same initial camera pose, and for each trajectory visualization, we additionally show the final camera pose at the end of the trajectory.
Our trajectory planning method yields more accurate and complete reconstructions, resulting in higher-quality renderings and effectively preventing holes or noise in the reconstructed surfaces.
}
\vspace*{0mm}
\label{fig:supp_qualitative_comparison_a}
\end{figure*}

\begin{figure*}

\setlength{\fboxsep}{0pt}   % padding
\setlength{\fboxrule}{1.pt}% border thickness

\vspace*{7mm}

\centering

% FisherRF
\begin{minipage}[b]{0.33\textwidth}
    \centering
    {\includegraphics[width=\linewidth]{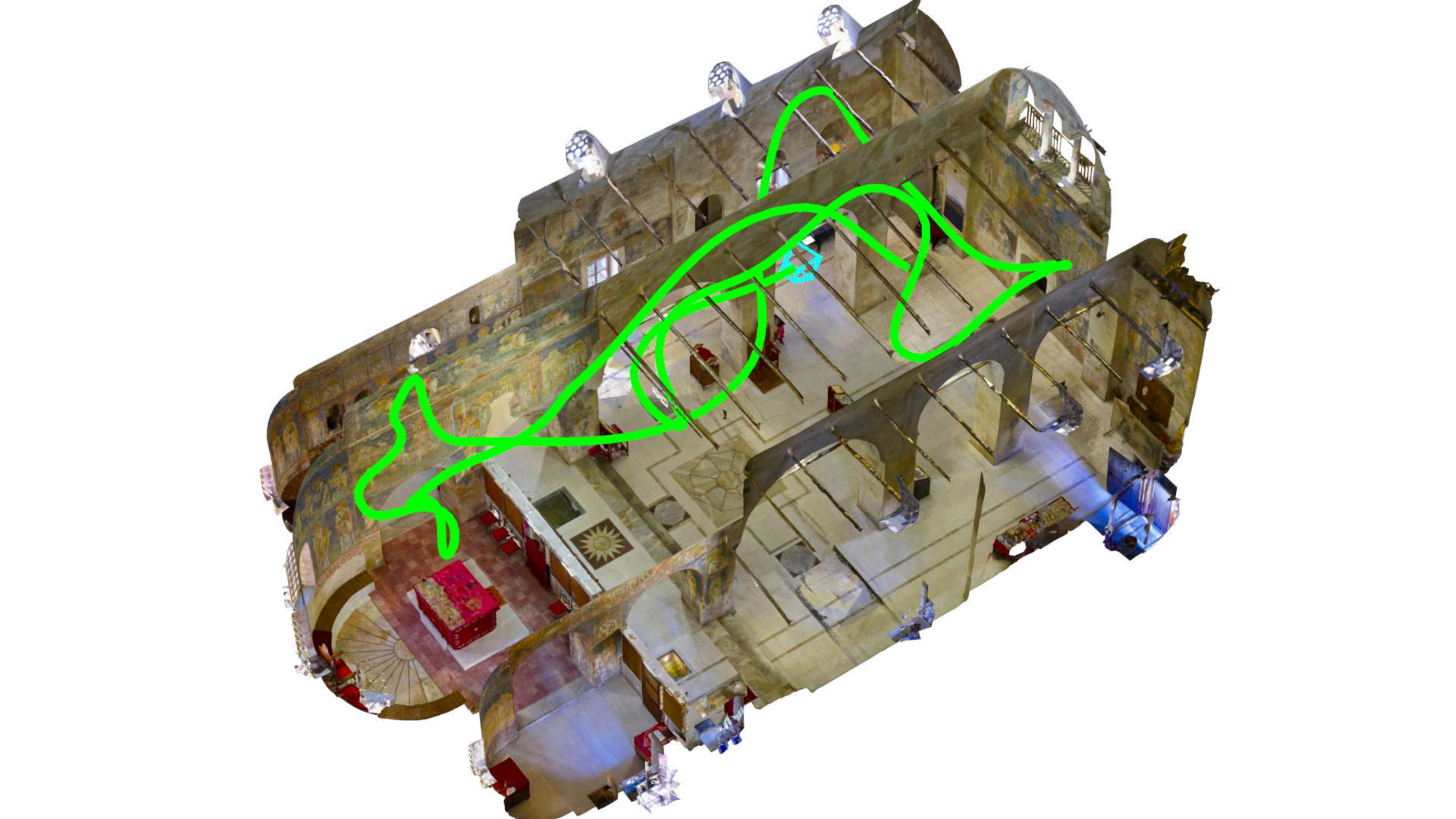}} 
    {\includegraphics[width=\linewidth]{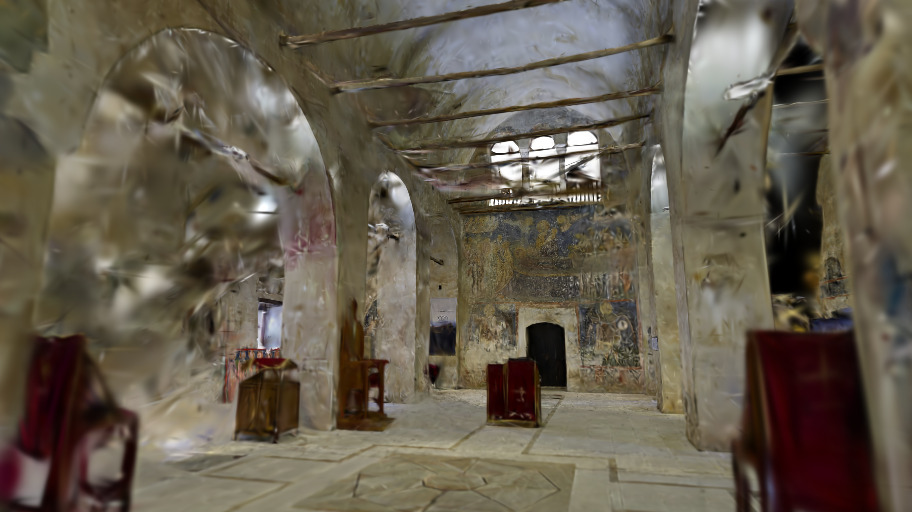}}
    {\includegraphics[width=\linewidth]{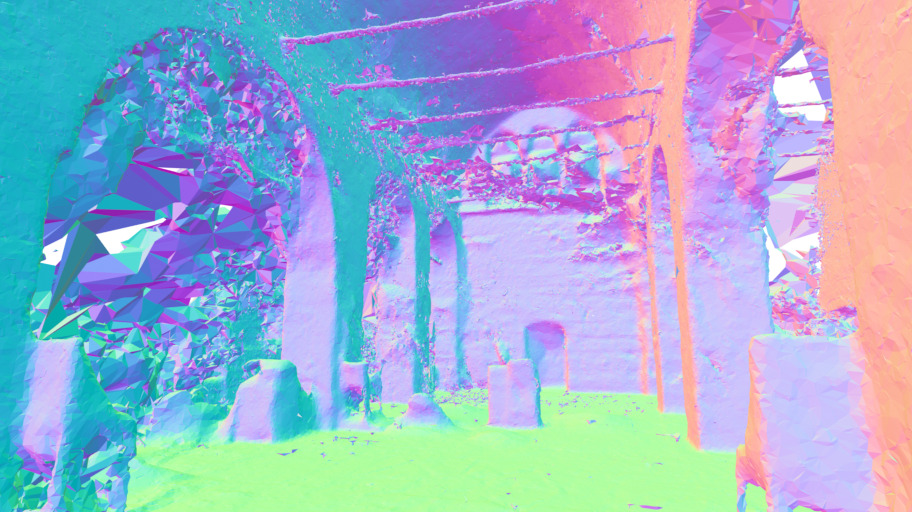}}
    {\includegraphics[width=\linewidth]{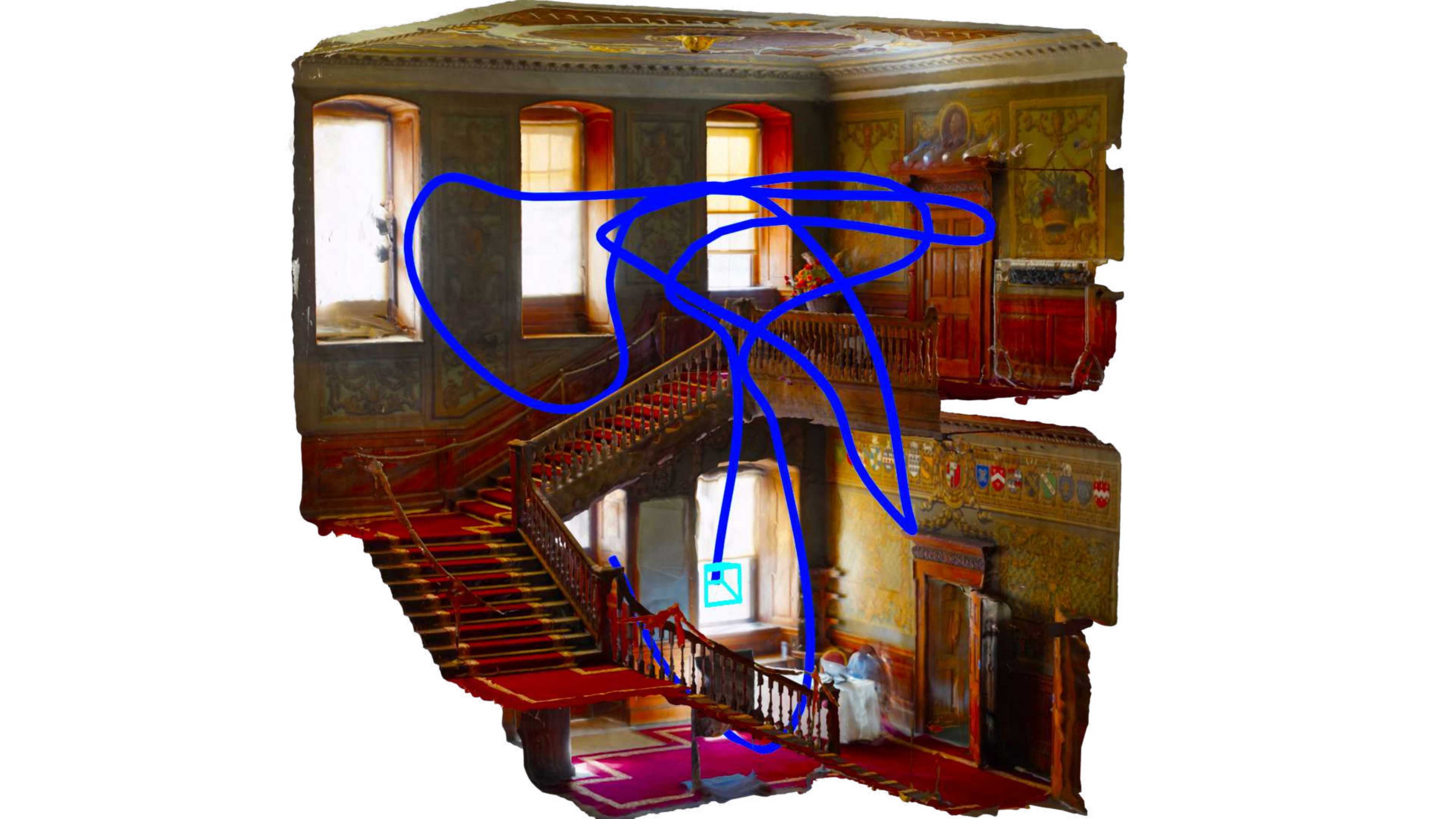}}
    {\includegraphics[width=\linewidth]{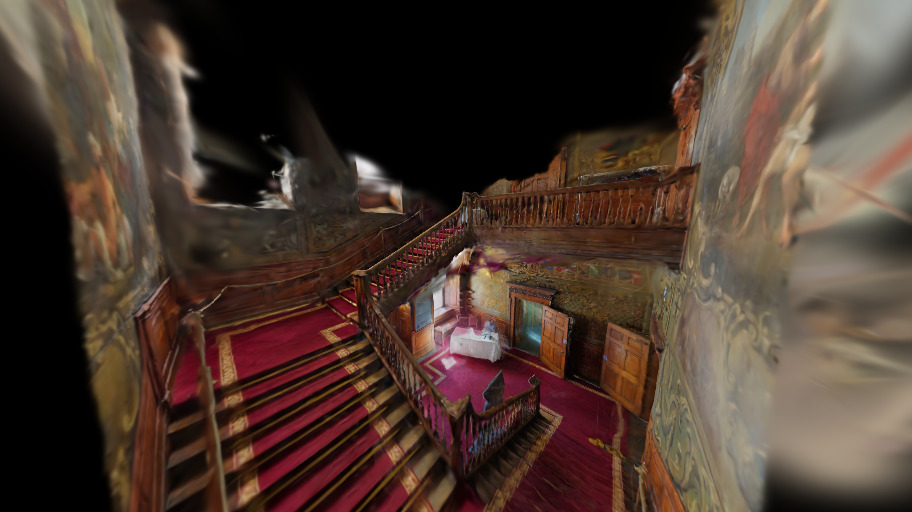}}
    {\includegraphics[width=\linewidth]{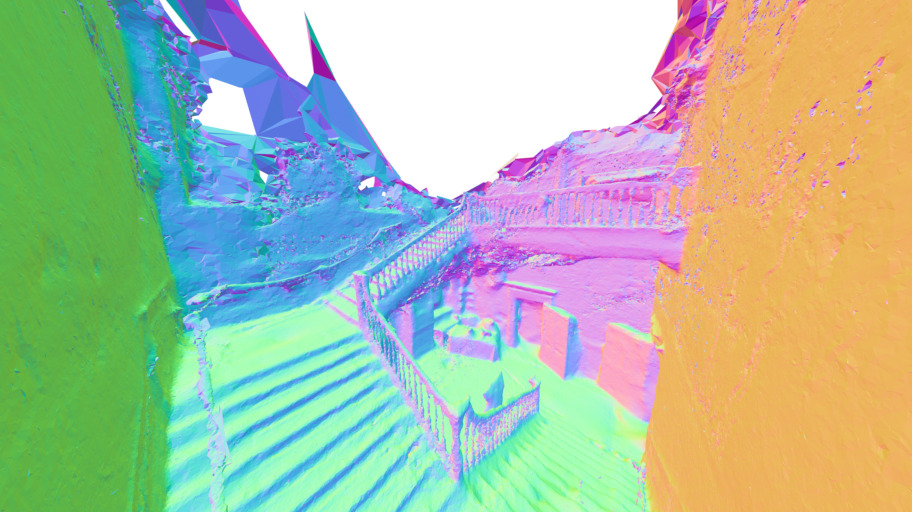}}
    \vfill
    FisherRF~\cite{jiang2024fisherrf}
\end{minipage}
\hfill
% MACARONS
\begin{minipage}[b]{0.33\textwidth}
    \centering
    {\includegraphics[width=\linewidth]{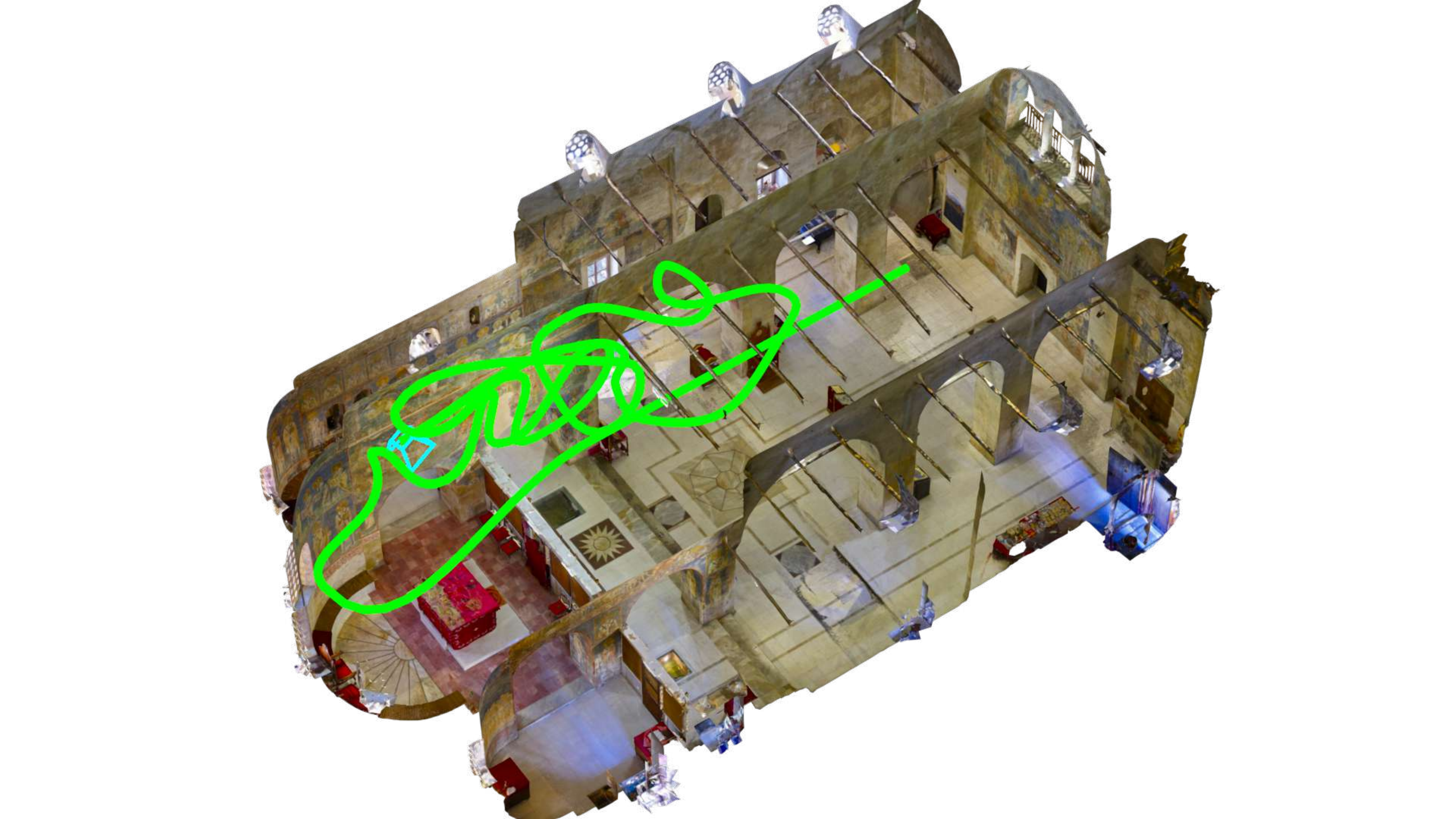}} 
    {\includegraphics[width=\linewidth]{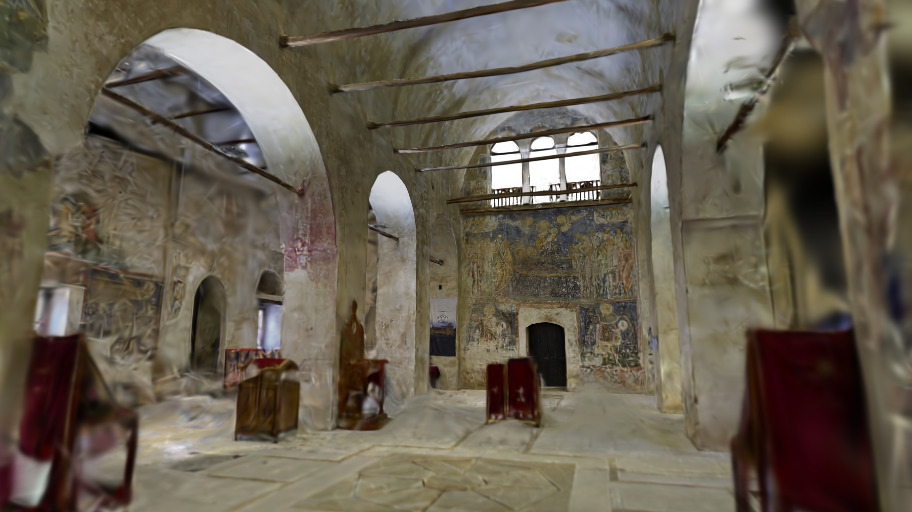}}
    {\includegraphics[width=\linewidth]{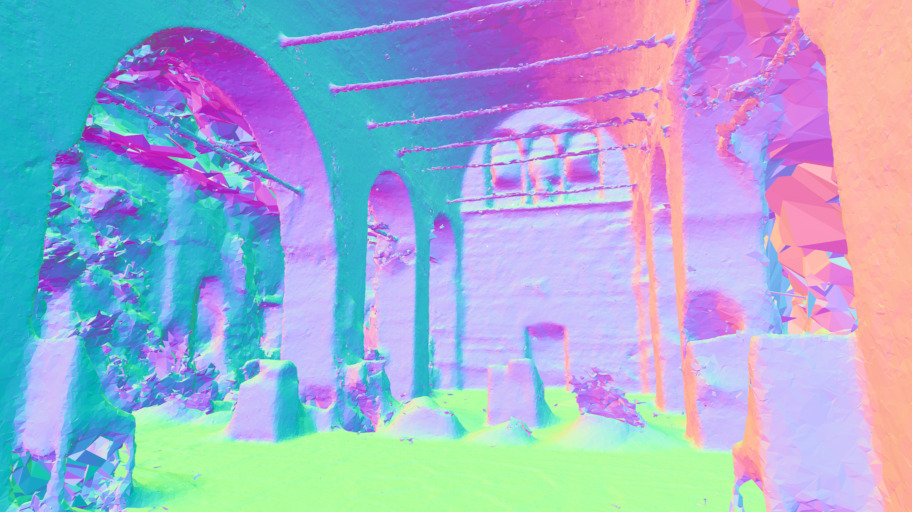}}
    {\includegraphics[width=\linewidth]{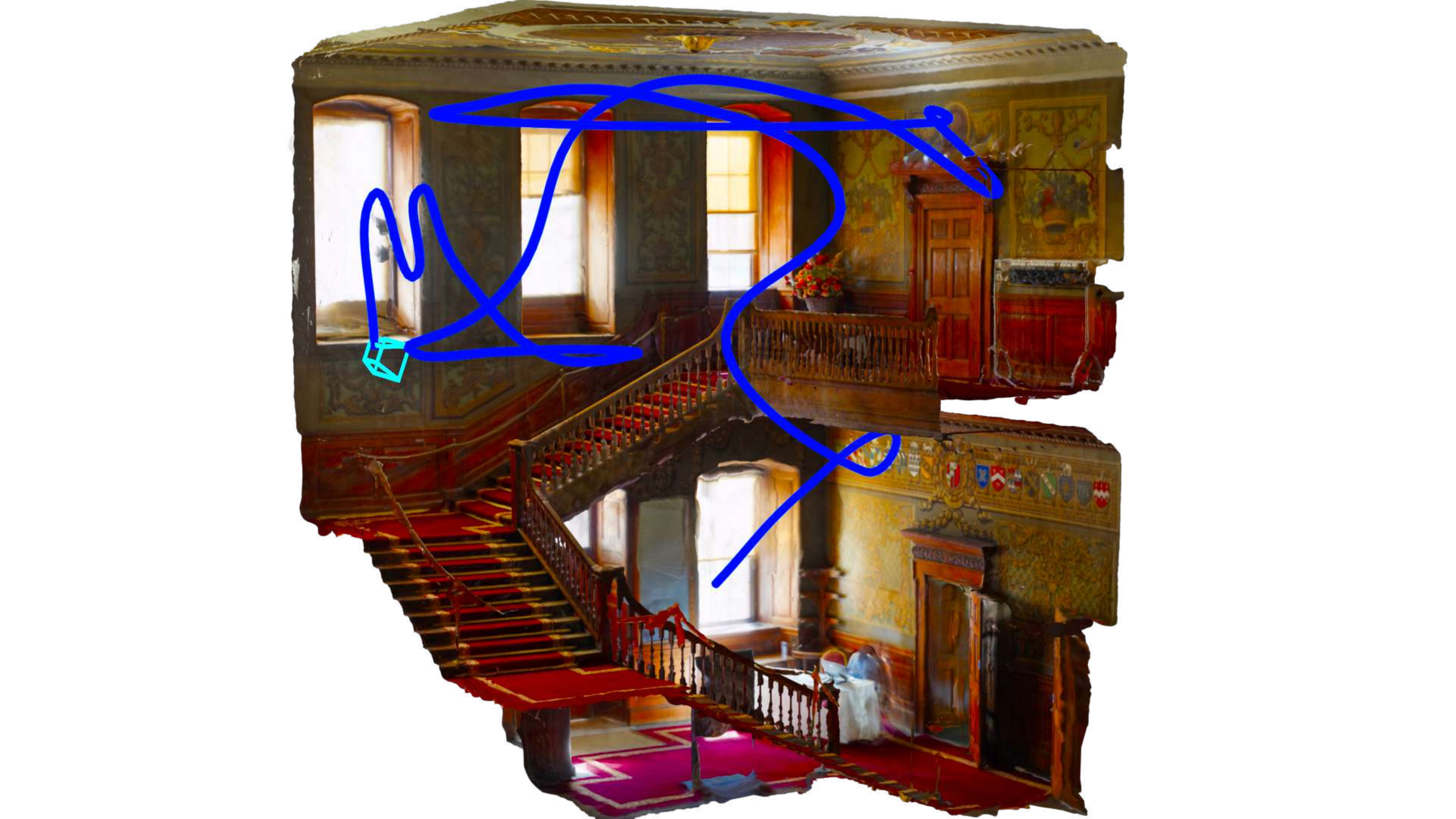}}
    {\includegraphics[width=\linewidth]{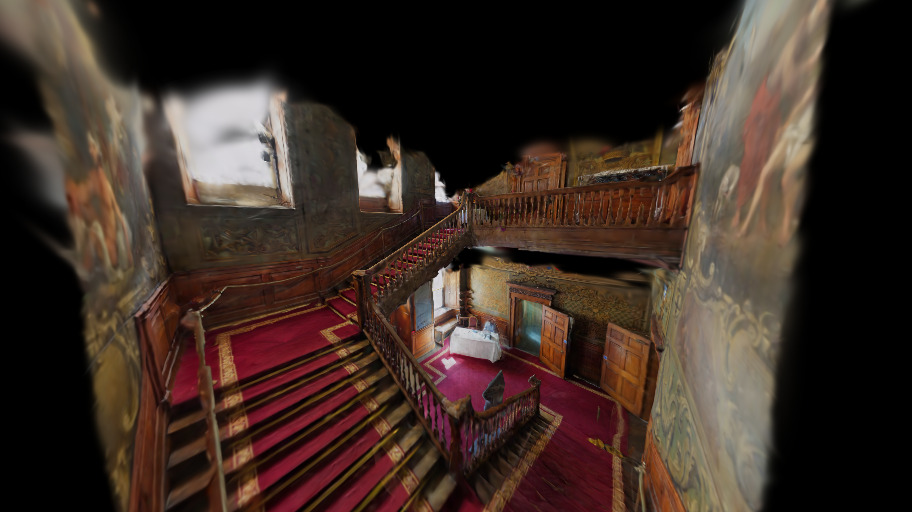}}
    {\includegraphics[width=\linewidth]{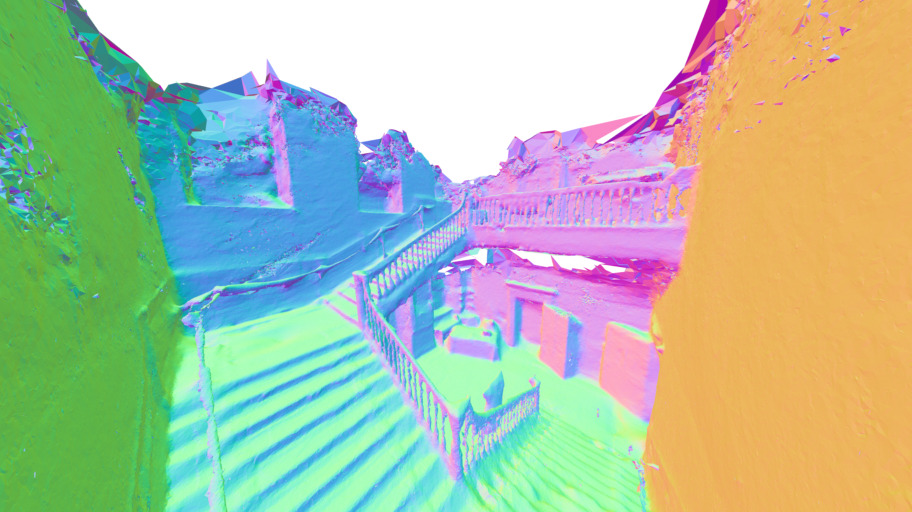}}
    \vfill
    MACARONS~\cite{guedon2023macarons}
\end{minipage}
\hfill
% Ours
\begin{minipage}[b]{0.33\textwidth}
    \centering
    
    {\includegraphics[width=\linewidth]{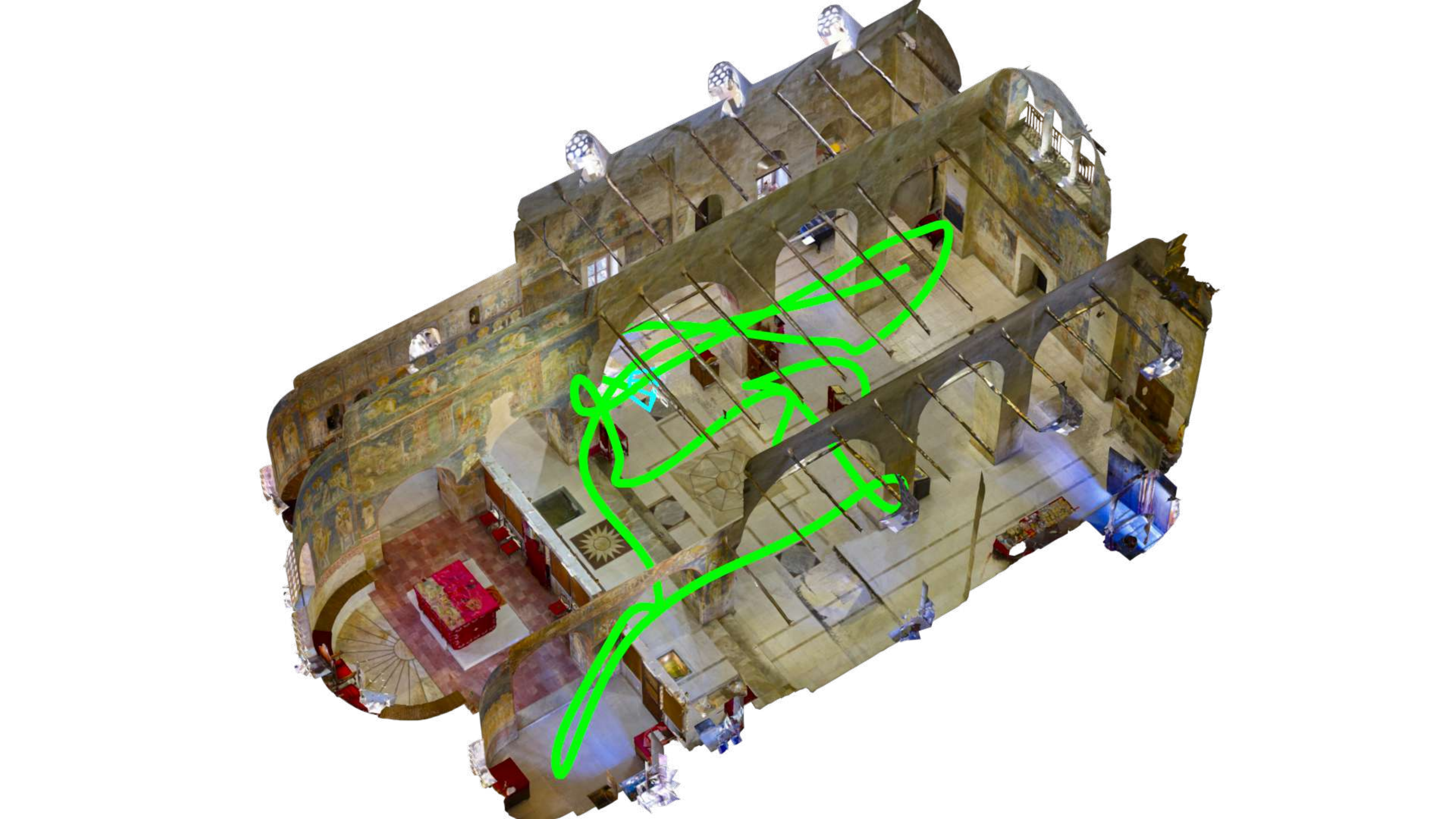}} 
    {\includegraphics[width=\linewidth]{sec/images/render/church_ours3_cam3/rgb.jpg}}
    {\includegraphics[width=\linewidth]{sec/images/render/church_ours3_cam3/mesh_normals.jpg}}
    {\includegraphics[width=\linewidth]{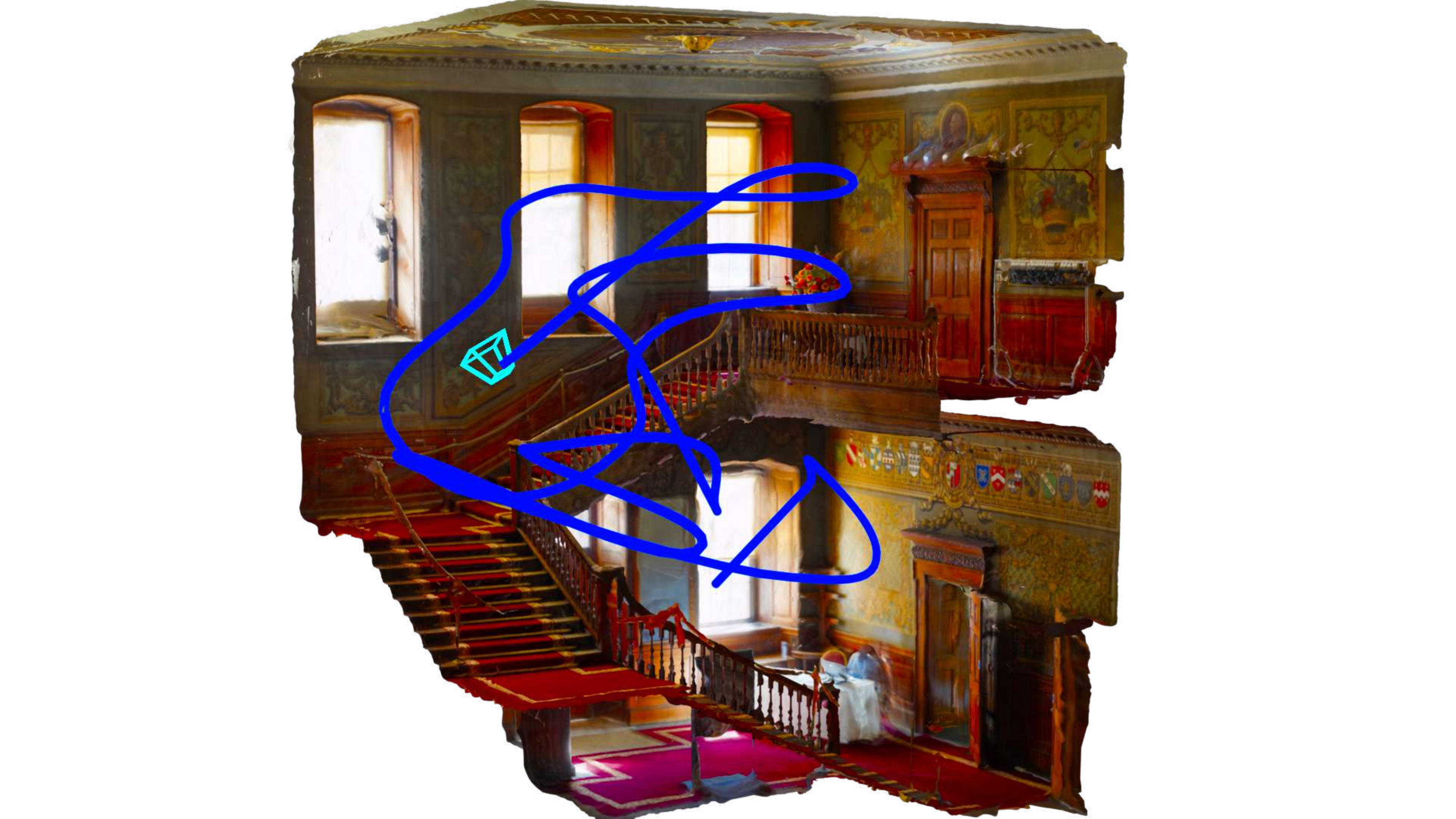}}
    {\includegraphics[width=\linewidth]{sec/images/render/stair_ours1_cam1/rgb.jpg}}
    {\includegraphics[width=\linewidth]{sec/images/render/stair_ours1_cam1/mesh_normal.jpg}}
    \vfill
    Ours
\end{minipage}

\caption{
% \vincent{
\textbf{Visualization of exploration trajectories and qualitative comparisons of novel view synthesis and surface reconstruction in indoor scenes. From top to bottom, the scenes are St. Sofia Church and Barts.} 
In the same scene, all methods start from the same initial camera pose, and for each trajectory visualization, we additionally show the final camera pose at the end of the trajectory.
Our trajectory planning method yields more accurate and complete reconstructions, resulting in higher-quality renderings and effectively preventing holes or noise in the reconstructed surfaces.
}
\vspace*{0mm}
\label{fig:supp_qualitative_comparison_b}
\end{figure*}

In \cref{fig:supp_qualitative_comparison_a} and \cref{fig:supp_qualitative_comparison_b}, we present visualizations of the exploration trajectories generated by different methods, where for each scene all methods start from the same initial pose, along with qualitative comparisons of novel view synthesis and mesh-based normal maps. Under an identical movement budget, our method achieves thorough exploration in both indoor and outdoor environments, resulting in high-quality reconstructions, whereas incomplete exploration by the other methods leads to noticeably inferior reconstruction quality.


\subsection{Additional Ablation Study}

\textbf{Impact of longer-range look-ahead steps $N_d$.} Table~\ref{tab:ablation_beams} presents the results for increased look-ahead steps $N_d>10$. Performance peaks when $N_d=15\text{--}20$; while it slightly declines for larger values, it remains superior to shorter look-ahead steps, as shown in \Cref{fig:heatmap}.

\setcounter{table}{7}
\begin{table}
\caption{Ablation study on look-ahead steps $N_d$.}
  \centering
  \setlength{\tabcolsep}{8pt}
  \scalebox{0.8}{
  \begin{tabular}{ccccccc}
    \toprule
    $N_d$ & 10 & 15 & 20 & 25 & 30 & 50 \\
    \midrule
    AUC  & 0.652 & \textbf{0.673} & 0.664 & 0.662 & 0.658 & 0.652\\
    Cov. & 0.888 & 0.887 & \textbf{0.892} & 0.879 & 0.881 & 0.878\\
    \bottomrule
  \end{tabular}
  }
  \label{tab:ablation_beams}
\end{table}

\noindent\textbf{Robustness under pose uncertainty.} We corrupt camera poses with Gaussian noise ($\sigma$ = 0.5m translation, 3° rotation) during planning. These are deliberately larger than typical localization errors to rigorously stress-test the method. Under this setting, performance decreases only marginally, with AUC dropping from 0.652 to 0.649 (-0.28 pp) and Cov. decreasing from 0.888 to 0.877 (-1.12 pp), demonstrating strong robustness to substantial pose uncertainty.

\noindent\textbf{Effect of proxy point sampling density.} Table~\ref{tab:density} shows that while increasing proxy point density leads to steady improvements in AUC and Cov. by refining coverage gain estimates, the performance remains relatively stable across a broad range of densities. This suggests that our method is robust to sampling density, with a $1\times$ density already providing a strong balance between estimation accuracy and computational overhead.

\begin{table}
  \centering
  \caption{Ablation study on proxy point sampling density ($1\times$ indicates the original density).}
  \setlength{\tabcolsep}{10pt} 
  \scalebox{0.81}{
  \begin{tabular}{ccccc}
    \toprule
    Density & $0.5\times$ & $1\times$ & $2\times$ & $4\times$ \\
    \midrule
    AUC  & 0.640 & 0.652 & 0.672 & \textbf{0.685} \\
    Cov. & 0.848 & 0.888 & 0.895 & \textbf{0.905} \\
    \bottomrule
  \end{tabular}
  }
  \label{tab:density}
\end{table}

\section{Failure Case and Analysis}
\label{sec:failure}

In a few scenes, we observe that the occupancy model exhibits reduced accuracy during the early stages of exploration, which leads to lower initial exploration efficiency. This limitation arises because the occupancy model is fundamentally geometric, relying on features extracted from local 3D neighborhoods. While such local geometric priors are effective at capturing generalizable primitives across scales and domains, they may be insufficient to provide a reliable global understanding when observations are sparse. As a result, the planner may not accurately identify the most informative regions at the beginning, leading to suboptimal estimation of coverage gain. However, as more observations are accumulated, the environment representation is progressively refined, and the system mitigates this issue through frequent closed-loop replanning, ultimately improving exploration performance over time.

\vspace*{\fill}
% \clearpage

% {
%     \small
%     \bibliographystyle{ieeenat_fullname}
%     \bibliography{main}
% }

\end{document}